\pdfoutput=1

\documentclass[11pt]{article}

\usepackage[final]{coling}

\usepackage{times}
\usepackage{subcaption}
\usepackage{latexsym}
\usepackage{color}

\usepackage[T1]{fontenc}

\usepackage[utf8]{inputenc}

\usepackage{microtype}

\usepackage{inconsolata}

\usepackage{graphicx}
\usepackage{amsmath}
\usepackage{bbm}
\usepackage{amsfonts}
\usepackage{booktabs}
\usepackage{multirow}
\usepackage{multicol}
\usepackage{listings}
\usepackage{caption}
\usepackage{color}
\usepackage[ruled,linesnumbered]{algorithm2e}

\newcommand{\app}{\raise.17ex\hbox{$\scriptstyle\sim$}}
\newcommand{\mytimes}{\raise.17ex\hbox{$\scriptstyle\times$}}
\newcommand{\myeq}{\raise.25ex\hbox{$\scriptstyle=$}}
\newcommand{\mylarge}{\raise.17ex\hbox{$\textstyle>$}}

\makeatletter
\def\UrlAlphabet{%
      \do\a\do\b\do\c\do\d\do\e\do\f\do\g\do\h\do\i\do\j%
      \do\k\do\l\do\m\do\n\do\o\do\p\do\q\do\r\do\s\do\t%
      \do\u\do\v\do\w\do\x\do\y\do\z\do\A\do\B\do\C\do\D%
      \do\E\do\F\do\G\do\H\do\I\do\J\do\K\do\L\do\M\do\N%
      \do\O\do\P\do\Q\do\R\do\S\do\T\do\U\do\V\do\W\do\X%
      \do\Y\do\Z}
\def\UrlDigits{\do\1\do\2\do\3\do\4\do\5\do\6\do\7\do\8\do\9\do\0}
\g@addto@macro{\UrlBreaks}{\UrlOrds}
\g@addto@macro{\UrlBreaks}{\UrlAlphabet}
\g@addto@macro{\UrlBreaks}{\UrlDigits}
\makeatother

\title{Momentum Posterior Regularization for Multi-hop Dense Retrieval }

\author{
 \textbf{Zehua Xia\footnotemark[1]\textsuperscript{1}},
 \textbf{Yuyang Wu\footnotemark[1]\textsuperscript{1,2}},
 \textbf{Yiyun Xia\textsuperscript{1,3}},
 \textbf{Cam-Tu Nguyen\footnotemark[2]\textsuperscript{1,3}}
\vspace{+.2cm}\\
 \textsuperscript{1}State Key Laboratory for Novel Software Technology, Nanjing University\\
 \textsuperscript{2}School of Computer Science, Nanjing University\\
 \textsuperscript{3}School of Artificial Intelligence, Nanjing University
\vspace{+.1cm}\\
 \small{
   \textbf{Emails:}
   \href{mailto:email@domain}{\{zehuaxia,wuyuyang,xiayiyun\}@smail.nju.edu.cn};
   \href{mailto:email@domain}{ncamtu@nju.edu.cn}
 }
}

\begin{document}
\maketitle
\renewcommand{\thefootnote}{\fnsymbol{footnote}} 
\footnotetext[1]{These authors contributed equally to this work and should be regarded as co-first authors.} 
\footnotetext[2]{Corresponding author.} 
\footnotetext[3]{Our code is available at \url{https://github.com/zeaver/mopo}}
\begin{abstract}

Multi-hop question answering (QA) often requires sequential retrieval (multi-hop retrieval), where each hop retrieves missing knowledge based on information from previous hops. To facilitate more effective retrieval, we aim to distill knowledge from a posterior retrieval, which has access to posterior information like an answer, into a prior retrieval used during inference when such information is unavailable. Unfortunately, current methods for knowledge distillation in one-time retrieval are ineffective for multi-hop QA due to two issues: 1) Posterior information is often defined as the response (i.e. the answer), which may not clearly connect to the query without intermediate retrieval; and 2) The large knowledge gap between prior and posterior retrievals makes existing distillation methods unstable, even resulting in performance loss. As such, we propose MoPo (Momentum Posterior Regularization) with two key innovations: 1) Posterior information of one hop is defined as a query-focus summary from the golden knowledge of the previous and current hops; 2) We develop an effective training strategy where the posterior retrieval is updated along with the prior retrieval via momentum moving average method, allowing smoother and effective distillation. Experiments on HotpotQA and StrategyQA demonstrate that MoPo outperforms existing baselines in both retrieval and downstream QA tasks\footnotemark[3].

\end{abstract}
\begin{figure*}
    \centering
    \includegraphics[width=0.99\linewidth]{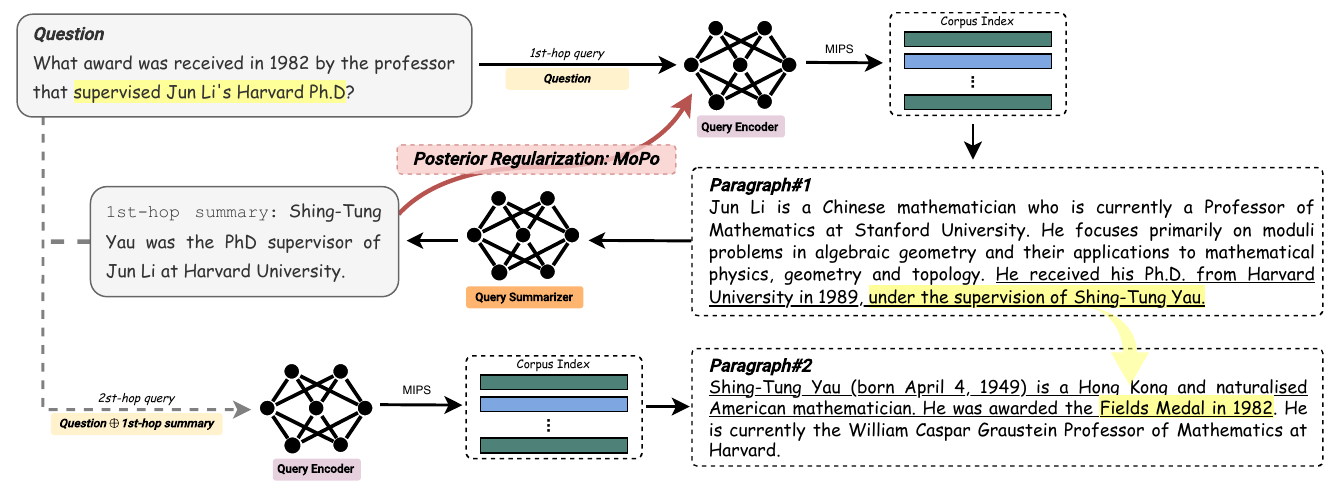}
    \caption{An example from HotpotQA benchmark. Given the 2-hop question, an iterative retriever is expected to retrieve the 1st and 2nd golden paragraph sequentially. After every retrieval step, a query-focused summary combining the query and the retrieved paragraph is generated, which is a kind of posterior information before conducting its retrieval. We utilize it to enhance the retriever training as shown by the red arrow. }
    \label{fig:fig1}
\end{figure*}
\section{Introduction}

LLMs demonstrate strong language capabilities, but their knowledge is not frequently updated, which makes it ineffective in responding to time-sensitive questions. In addition, LLMs still suffer from hallucinations, especially in knowledge-intensive question-answering tasks. Although augmenting LLMs with retrieval is a promising solution, a single round of retrieval is often insufficient for complex, multi-hop queries \citep{yang2018hotpotqa}. Recent approaches have focused on multi-hop reasoning with LLM  \citep{trivedi-etal-2023-ircot,zhang2024endtoendbeamretrievalmultihop}, multi-hop reranking \cite{zhang2024endtoendbeamretrievalmultihop}, or multi-hop dense retrieval \cite{xiong2021answeringcomplexopendomainquestions}. The first two approaches, when being used without a retrieval, are less efficient than multi-hop dense retrieval. This is because they require costly attention computation between the query and documents during inference. In contrast, multi-hop dense retrieval separately encodes queries and documents, enabling the precomputation of document embeddings for offline indexing and efficient search. 

Inspired by the success of distilling posterior information (e.g. answers) to one-time retrieval \citep{chen-etal-2020-bridging,Posterior-GAN}, we target an unexplored research question of \textit{``how can we effectively distill posterior information to facilitate multi-hop dense retrieval in complex, multi-hop QA?''} Our intuition is that such posterior information and the query can play as ``anchor information,'' thus reducing the ``semantic shift'' issue \cite{xiong2021answeringcomplexopendomainquestions} and facilitating multi-hop retrieval. 
Unfortunately, two main challenges hinder the adoption of existing methods in multi-hop QA. Firstly, most previous methods treat responses as the posterior information to train the posterior retrieval. 
However, in multi-hop QA, the response (answer) is not directly related to the original query. 
As a result, the posterior retrieval trained on such information may misguide the prior retrieval, consequently missing the important information at the intermediate hops. 
Secondly, the knowledge gap between prior and posterior retrievals makes it difficult for knowledge distillation. Although this issue exists in one-time retrieval, the gap is generally larger in multi-hop retrieval. Our empirical results show that, even with a suitable posterior information, the existing methods \cite{chen-etal-2020-bridging} do not work well for multi-hop retrieval, even resulting in a drop in retrieval performance. 

To address the aforementioned issues, this study incorporates posterior information to multi-hop dense retrieval with two main ideas. Firstly, rather than defining the final answer as the posterior information, we employ query-focused summary of the gold knowledge in the previous and the current hops as the posterior information for the current hop (see Figure \ref{fig:fig1}). By doing so, we ensure a stronger correlation between the posterior information and the current hop context. To facilitate training, we develop PostSumQA, a dataset derived from HotpotQA with $22,696$ questions. In PostSumQA, we provide posterior summary annotation at every hop for all multi-hop questions. Secondly, we propose \textit{Momentum Posterior Regularization} (MoPo), which exhibits smoother convergence and is easier to train compared to existing posterior regularization \cite{chen-etal-2020-bridging}. In MoPo, the posterior model is updated at each training step using a momentum-based moving average of the prior model, thus reducing the knowledge gap between the prior and posterior retrieval models. Experimental results on HotpotQA and StrategyQA show that MoPo effectively exploits posterior information, resulting in better retrieval performance compared to previous multi-hop retrieval methods and existing posterior regularization methods. When being used in the traditional retrieval-reranking-generation, MoPo helps improve the downstream tasks, the reranking and QA tasks. Particularly, a simple pipeline with MoPo as the retrieval outperforms contemporary methods based on multi-hop reranking and multi-hop reasoning with LLM.

Our contributions can be summarized as follows:
    \begin{itemize}
        \item To our knowledge, we are the first to introduce posterior query-focused summary for multi-hop dense retrieval. Toward this end, we present PostSumQA, a high-quality dataset comprising 22,696 entries in English, tailored for training models on posterior information.
        \item This study proposes Momentum Posterior Regularization (MoPo), a simple yet effective posterior regularization framework for multi-hop dense retrieval. 
        \item Our method is empirically tested on the HotpotQA and StrategyQA datasets, outperforming recent baselines both in retrieval and downstream tasks (reranking and QA).
    \end{itemize}

\section{Related Works} 
\label{related work}
\paragraph{Multi-hop Dense Retrieval} 
Recent studies extend the dense retrieval framework \cite{devlin2019bertpretrainingdeepbidirectional} to support multi-hop QA. 
MDR \cite{qi2021answeringopendomainquestionsvarying} proposes an iterative framework, where the retrieval of each hop depends on the previous retrieved documents in the previous steps. Here, MDR exploits a dual-encoder \cite{related_dpr_2020} as the single-time retrieval for each hop retrieval. 
BeamDR \cite{zhao2021multistepreasoningunstructuredtext} is similar to MDR in that it aims to retrieve the next document depending on the document candidates in the beam. Its optimization objective, however, is based on contrasting positive retrieval chain to the negative retrieval chain, rather than contrasting hop-level samples as in MDR. 
Our research builds on these studies, yet we aim to incorporate posterior information into the retrieval process.

\paragraph{Multi-hop Reranking} 
It is common to exploit an inefficient but effective cross-encoder to rerank the retrieved candidates \cite{xiong2021answeringcomplexopendomainquestions} for better knowledge selection. The cross-encoder model requires the expensive cross-attention between a query and a document at the inference time; thus, is only efficiently used with a small set of candidates, typically from an efficient retrieval. Recently, \citet{ma2023chainofskillsconfigurablemodelopendomain} proposes Chain-of-Skills (CoS) that adopts a similar framework with MDR but performs reranking after every hop. Unlike CoS, which interleaves between retrieval and reranking, BeamRetrieval \cite{zhao2021multistepreasoningunstructuredtext} performs direct multi-hop reranking from a set of document candidates, that is obtained in advance.
 
\paragraph{Multi-hop Reasoning with LLM} 
Thanks to the strong reasoning capability of LLM \cite{cot_wei_nips}, recent studies treat LLM as a sophisticated retrieval-generation agent for multi-hop QA. Representative works being SelfAsk \cite{press-etal-2023-selfask}, IRCoT \cite{trivedi-etal-2023-ircot}, FLARE \cite{jiang-etal-2023-flare}, and BeamAggR \cite{chu-etal-2024-beamaggr}.
In general, these methods combine LLM and retrieval in two ways: 1) Performing question decomposition with LLM and using a retrieval to help generate $K$ answers for each simple question in the question tree \cite{chu-etal-2024-beamaggr}; 2) Interleaving between query reformalization and retrieval \cite{press-etal-2023-selfask, trivedi-etal-2023-ircot, jiang-etal-2023-flare}. The latter bears some resemblance to our framework, yet we focus on improving the retrieval with posterior information. It is noteworthy that these methods are much more costly than MoPo due to the use of resource-intensive LLM for reranking, query formalization, and answer generation. 

\paragraph{Posterior Knowledge Enhancement} Posterior knowledge has been used to refine knowledge selection in dialogue systems using \cite{kim2020sequentiallatentknowledgeselection, chen-etal-2020-bridging}. In general, these methods aim to perform knowledge distillation between the posterior and the prior retrievals by minimizing the KL divergence.  Unlike these methods, however, we focus on mult-hop dense retrieval setting.

\section{Methodology}
\subsection{Problem Definition}
For open-domain multi-hop question answering retrieval: given the question $q$ and a large textual corpus $\mathcal{D}$, a retriever need to retrieve a sequence $\mathbf{D}_{seq} = \{d_1,\ldots,d_L\}$ of $L$ relevant documents to construct the reasoning chain and finally find the target answer $a$.
In practice, the retriever returns the $K$ documents with the highest scores as candidates for downstream modules, like reranker or reader/generation, where $|\mathcal{D}| \gg K$.

\subsection{Iterative Multi-hop Dense Retriever}
Inspired by MDR \cite{xiong2021answeringcomplexopendomainquestions}, we model the probability of a sequence of documents given the query based on the dense retrieval model $\mathbf{M}_{\theta}$ with the parameters $\theta$:
\begin{align}
    P_{\theta}(\mathbf{D}_{seq}|q) = \prod_{t=1}^{L} P_{\theta}(d_t|q,d_1, ..., d_{t-1}) \label{eq:retrieval_1}
\end{align}
\noindent where, $d_t$ represents the retrieved document at step $t$, and when $t=1$, query is the original question.
After finishing $t-$th retrieval, we apply some post-processing to $q_t$ for next time.
We define $G_{s}$ as the query post-processing module:
\begin{align}
     q_t = G_{s}(q_{t-1}, d_{t-1})
\end{align}
Accordingly, the Equation (\ref{eq:retrieval_1}) is simplified as:
\begin{align}
     P_{\theta}(\mathbf{D}_{seq}|q) = \prod_{t=1}^{L} P_{\theta}(d_t|q_t)\label{eq:retrieval_2}
\end{align}
Then the InfoNCE contrastive loss \cite{oord2019representationlearningcontrastivepredictive} function for a tuple within a batch $ (q, \mathbf{D}_{seq}) \sim \mathbf{B}$ is as follows 
\begin{align}
    &\mathcal{L}_{\operatorname{InfoNCE}}(\theta,\mathbf{B}) \label{eq:infonce_1} \\
    &\;\;\;\;\;\;=\mathbb{E}_{r\sim\mathbf{B}}\Big[\sum_{t=1}^L-\log\frac{\operatorname{exp}\left({f_{\theta} (q_t,d_t^{+})}\right)}{\sum_{d\in d_t^{\pm}} \operatorname{exp}\left( f_{\theta}\left( q_t, d\right) \right)} \Big]\nonumber
\end{align}
with $r=(q, \mathbf{D}_{seq})$ where $q$ is the original question,  $q_t$, $d^+_t$ and $d^{-}_t$ represent the query at the t-hop and the corresponding positive ($d^+_t\in\mathbf{D}_{seq}$) and negative documents at t-th hop, respectively. Here, $f_{\theta}(\cdot)$ indicates the similarity function, which exploits $\mathbf{M}_{\theta}$ to map the query and the document into two dense vectors for similarity measurement like Exact Inner Product in MoPo.

\subsection{Posterior Summary Utilization}
We introduce posterior summary to reformulate query and enhance the retrieval capability of $\mathbf{M}_{\theta}$.

\paragraph{Query Reformulation} 
In MDR, the post-processing module $G_s$ performs a simple concatenation of the query and retrieved documents. However, such a method will cause an increment in the query length. On the other hand, if $G_s$ is just a simple summarization, the semantic drift issue is highly likely to occur.
As a result, we retain the original question $q$, and concatenate the summary $s_{t-1}$ generated from retrieved documents as the query for the $t-$th step retrieval:
\begin{align}
     q_t = q \oplus s_{t-1} = q \oplus G_{s}(s_{t-2}, d_{t-1}, q) 
\end{align}
The symbol $\oplus$ represents concatenation operation.
Intuitively, the syntactic and semantic consistency across retrieval steps can be ensured in this way.
For syntactic consistency, the query length will not increase significantly thanks to the summarization operation. In addition, except for the first-hop retrieval step, subsequent queries all have the same structure --- the original question and the summary of all previously retrieved documents.
For semantic consistency, the original question is included at every step, subsequently mitigating the semantic drift issue. 
Equation (\ref{eq:retrieval_2}) is then rewritten as:
\begin{align}
    P(\mathbf{D}_{seq}|q) &= \prod_{t=1}^{L} P(d_t|s_{t-1}, d_{t-1}, q)
\end{align}

\paragraph{Posterior Summary Enhanced Retriever} 
It is intuitive that a retrieval that exploits both the question and the answer (the posterior information) is more effective than a retrieval that only uses a question for the search \cite{chen-etal-2020-bridging}. In this paper, we define the posterior information for each hop (step) as the query-aware summary up to that step.
Specifically, for the retrieval step $t$, as the search query $q_t$ is formed by the original query $q$ and the previous step summary $s_{t-1}$,  the summary $s_t$ is the posterior information for this particular step.
Here, we denote the posterior-enhanced retriever with parameters $\phi$ by $\mathbf{M}_{\phi}$, which has the same architecture and similarity function with $\mathbf{M}_{\theta}$ for simplicity.
Given a tuple $(s_{t}, d_{t}^{+}, \mathbf{d}^{-}_{t})$, similar in Equation (\ref{eq:infonce_1}), the posterior similarity during training is:
\begin{align}
    p_{\phi}(d_{t}|s_{t}) = \frac{\operatorname{exp}\left({f_{\phi} (s_{t},d^{+}_{t})}\right)}{\sum_{d\in d^{\pm}_{t}} \operatorname{exp}\left( f_{\phi}\left( s_{t}, d\right) \right)}\label{eq:posterior_t}
\end{align}
To train $\mathbf{M}_{\theta}$,
we sample a tuple $(q, \mathbf{S}_{seq}, \mathbf{D}_{seq}) \sim \mathbf{B}^{'}$, where $\mathbf{S}_{seq}$ indicates the query-focused summary sequence. We then calculate the posterior regularized InfoNCE loss as follows:
\begin{align}
    &\mathcal{L}(\theta, \mathbf{B}^{'}) = \mathcal{L}_{\operatorname{InfoNCE}}(\theta,\mathbf{B}^{'}) \label{eq:pr}\\
    &\;\;\;\;\;\;+ \lambda\cdot \mathbb{E}_{r^{'}\sim \mathbf{B}^{'}}\big[ \sum_{t=1}^L D_{\operatorname{KL}}(p_{\phi}(d_t|s_t)\|p_{\theta}(d_t|q_t)) \big]\nonumber 
\end{align}
with $r^{'}=(q, \mathbf{S}_{seq}, \mathbf{D}_{seq})$, and $\lambda$ is the Kullback–Leibler (KL) divergence loss weight.

\subsection{Momentum Posterior Regularization}\label{sec:more}

    





\begin{algorithm*}[h]
\KwIn{Momentum coefficient $m$; Prior model $\mathbf{M}_{\theta}$}
\KwData{Training dataset $\mathbf{X}$}
$\mathbf{M}_{\phi} \leftarrow \mathbf{M}_{\theta}$\tcp*{Initialize posterior model $\mathbf{M}_{\phi}$}
\For{$(q, \mathbf{D}_{seq}, \mathbf{S}_{seq})$ \text{\textbf{in}} $\mathbf{X}$}{
    $P_{\theta}(\mathbf{D}_{seq}|q)\leftarrow M_{\theta}(q, \mathbf{D}_{seq})$\tcp*{Compute prior logit in Equation (\ref{eq:infonce_1})}
    $\phi \leftarrow m\phi + (1-m)\theta$\tcp*{Momentum update $\mathbf{M}_{\phi}$ in Equation (\ref{eq:mopo})}
    $P_{\phi}(\mathbf{D}_{seq}|\mathbf{S}_{seq})\leftarrow M_{\phi}(q, \mathbf{S}_{seq},\mathbf{D}_{seq})$\tcp*{Compute posterior logit in Equation (\ref{eq:posterior_t})}
    $\mathcal{L}(\theta) \leftarrow \mathcal{L}_{\operatorname{InfoNCE}}\left[P_{\theta}(\mathbf{D}_{seq}|q)\right] + \lambda\cdot\operatorname{KL}\left[P_{\phi}(\mathbf{D}_{seq}|\mathbf{S}_{seq})\|P_{\theta}(\mathbf{D}_{seq}|q)\right]$\tcp*{Equation (\ref{eq:pr})}
    Update $M_{\theta}$ with $ \mathcal{L}(\theta)$ by Adam optimizer \cite{adam_2015};
}
\caption{Momentum Posterior Regularization}\label{alg:code}
\end{algorithm*}
Usually, we adopt a two-stage training strategy for Posterior Regularization (PR): firstly train a posterior model $\mathbf{M}_{\phi}$ on the whole training tuples $\{(\mathbf{S}_{seq}, \mathbf{D}_{seq}, \mathbf{d}^{-})\}$, and then employ $\mathbf{M}_{\phi}$ to compute the KL divergence with the retriever $\mathbf{M}_{\theta}$ training on $\{(q, \mathbf{S}_{seq}, \mathbf{D}_{seq})\}$. 
However, this solution yields poor results on retrieval evaluation.

\paragraph{Analysis of PR Loss} We analyze the training loss to understand the failure cause of 2-stage PR strategy.
In the first stage, during the training of the posterior model $\mathbf{M}_{\phi}$, we observe that the training loss converges rapidly.
In contrast, in the second stage, which involves training the prior model $\mathbf{M}_{\theta}$, both the InfoNCE and Kullback-Leibler (KL) losses exhibit slow convergence rates. This is particularly pronounced for the KL term.
Moreover, we note that the KL term consistently accounts for a high proportion of the total loss, even when we reduce its weight $\lambda$. 
This behavior is unexpected and potentially problematic, as the KL term is intended to serve as a regularization component rather than dominate the training process.

\paragraph{Momentum Update}
We hypothesize that this suboptimality may be attributed to the overly influential supervised training signal from the posterior summary.
For this reason, it is desirable that the posterior regularization is smooth and not overly strong.
Therefore, we propose a momentum posterior regularization framework to address this issue.
At training step $\tau$, given the momentum coefficient $m$, we update $\phi$ by:
\begin{align} 
    \phi^{(\tau)} \leftarrow m\phi^{(\tau-1)} + (1-m)\theta^{(\tau-1)}
\label{eq:mopo}
\end{align}
The posterior distribution is obtained only by passing the query with posterior information ($s_t$) through $M_\phi$ in forward-passing. In essence, only the parameters of prior model $\theta$ are updated by back-propagation. By doing so, it simplifies the original two-stage training for posterior regularization to one-stage training strategy, making the training simpler compared to PR.

\section{Data Preparation with Backward Summary Generation}
\label{Data Synthesis}

MoPo training requires a dataset with posterior information annotation, i.e. query-focused summaries. In this paper, we aim at an automatic method to construct such a dataset from a multi-hop QA dataset. More specifically, given a sequence of $(q, d_1, d_2, ..., d_L, a)$, where $q, a$ are the question, answer and $(d_1, d_2, ..., d_L)$ is the document sequence containing key information to derive the answer, the objective is to generate $(q, \ldots, d_t, s_t, \ldots, a)$, where $s_t$ is the summary at the t-th hop. The direct intuition is to exploit LLM to help generate such summaries in the forward direction. In other words, we can aim to generate $s_{t}$ given $q, d_t, s_{t-1}$. However, doing so can lead to semantic drift as LLM may include redundant information from $d_t$, for it does not know which key information is needed for the final answer. 

In this paper, we propose a \textit{backward summary generation} for data generation. Specifically, for the last hop (the $L$-th hop), we use the rule-based method QA2D \cite{demszky2018transformingquestionansweringdatasets} to generate $s_L$ from $q$ and $a$, making sure that there is no redundant information in $d_L$ included in $s_L$. For the intermediate $t$-th step, given $s_{t+1}$ summary in the next hop, the query $q$, and the current hop document $d_t$, we ask LLM to generate $s_t$ from $q, d_t, s_{t+1}$ by removing redundant information in $s_{t+1}$ that is not included in  $d_t$. Our experience shows that having access to the ``look ahead'' information in $s_{t+1}$ facilitates the hop summary generation. Experiments in Appendix \ref{sec:Data Synthesis} show the advantage of backward summary generation in generating high-quality data compared to the forward summary process. By applying the above process to a subset of the HotpotQA training set \cite{yang2018hotpotqa}, a multi-hop QA set in English, we obtain 22,696 data points for MoPo training. We refer to this dataset as PostSumQA and publish it for future research. Detailed information and analysis of this dataset are provided in the Appendix.

\section{Retrieval Experiments}
\label{retrieval_experiment}
\begin{table*}[t]
\centering
\scalebox{1}{
\begin{tabular}{p{2cm}cccccccc@{}}
\toprule
\textbf{Model}                & R@2   & R@20    & R@50      & R@100 & EM@2   & EM@20 & EM@50      & EM@100 \\ \midrule
\multicolumn{9}{c}{HotpotQA}  \\ \midrule
BeamDR & - & -  & -   & 92.90  & \underline{60.70} & - & - & 79.20\\
MDR$\rm{}_{origin}$ & 65.90 & 80.20  & -   & -  & - & - & - & -\\
MDR$\rm{}_{zero}$ & 92.12 & 92.64  & 93.77   & 94.62  & 46.11 & 58.27 & 69.37 & 73.09\\
MDR  & 94.34 & 94.85  & 95.63   & 96.38   & 55.96 & 71.73 & 76.82 & 79.70     \\
MDR$\rm{}_{sum}$       & \underline{94.66} & \underline{95.27}  & 95.85   & 96.38  & 60.04 & 72.65 & 76.98 & 79.94     \\
PR$\rm{}_{fixed}$ & 94.49 & 95.07  & \underline{95.88}   & 96.55  & 57.67 & 72.18 & 76.93 & 79.95     \\
PR$\rm{}_{dyn}$ & 94.52 & 95.13  & 95.69   & \underline{96.61}  & 57.33 & \underline{72.67} & \underline{77.13} & \underline{80.17}     \\
MoPo         & \textbf{94.77} & \textbf{95.43}  & \textbf{96.27}   & \textbf{96.70} & \textbf{63.03} & \textbf{76.74}  & \textbf{80.27}   & \textbf{82.20}     \\
\midrule
\multicolumn{9}{c}{StrategyQA}  \\ \midrule
MDR$\rm{}_{zero}$ & 42.80 & 43.15  & 43.52   & 43.85  & 27.96 & 32.44 & 35.83 & 37.34\\
MDR  & 42.64 & 42.85  & 43.21   & 43.70   & 25.31 & 31.87 & 35.29 & 36.06     \\
MDR$\rm{}_{sum}$       & \underline{42.85} & \underline{43.31}  & \underline{43.66}   & \underline{43.87}  & \underline{28.88} & \underline{32.92} & \underline{36.46} & \underline{37.40}     \\
PR$\rm{}_{fixed}$ & 42.38 & 42.92  & 43.39   & 43.80  & 25.76 & 32.31 & 36.05 & 36.53     \\
PR$\rm{}_{dyn}$ & 42.80 & 43.14  & 43.47   & 43.78  & 25.81 & 32.84 & 36.14 & 37.14     \\
MoPo         & \textbf{43.36} & \textbf{43.61}  & \textbf{43.94}   & \textbf{44.15} & \textbf{31.91} & \textbf{35.61}  & \textbf{37.90}   & \textbf{39.24}     \\
\bottomrule
\end{tabular}}
\caption{Retrieval performance in recall and EM at k retrieved passages within $10\mytimes20$ search space. Results of MDR$\rm{}_{origin}$ and BeamDR come form their own paper. We also evaluate retrieval performance in different base models in Table \ref{retrieval contriever} and Table \ref{retrieval zero-shot}.}
\label{retrieval result}
\end{table*}

\subsection{Experimental Setup}
\paragraph{Datasets}
We test the retrieval and comprehensive reasoning ability of MoPo on two datasets: HotpotQA \cite{yang2018hotpotqa} and StrategyQA \cite{geva2021strategyqa}. 
HotpotQA is a multi-hop question answering dataset in the open domain, necessitating information from two separate Wikipedia pages to respond to a query.
It includes $113$K multi-hop questions and $\app$5M documents.
Both dev/test sets of HotpotQA have $\app$7K samples.
StrategyQA is another open-domain multi-hop question answering dataset with 2,780 examples and $\app$36M documents, where the reasoning process is not explicitly stated in the question, requiring 2 hops of information retrieval and strategic thinking to derive the answer. On StrategyQA, we directly utilize the model finetuned on HotpotQA for evaluation. As a result, we consider StrategyQA is a held-out dataset that allows us to the generalization of compared methods. 

\paragraph{Metrics}
Like \cite{xiong2021answeringcomplexopendomainquestions}, we use Recall and Exact Match (EM) metrics to evaluate the performance. Retrieval EM measures the percentage of test queries of which \textit{at least one of the retrieved sequences exactly matches that of the golden document sequence}. On the other hand, Recall measures the percentage of test queries of which at least one of the retrieved sequences containing \textit{at least one document of the golden document sequence}. More detailed information can be found in Appendix \ref{metric_detail}. As multi-hop QA requires information from all the hops to get the answer, EM is a more important metric for multi-hop retrieval.
Additionally, the beam size, or the number of candidate documents to achieve at each hop, is an essential parameter for final retrieval performance.
We follow the golden passages order of MDR \cite{xiong2021answeringcomplexopendomainquestions} but reduce the beam size from $50\mytimes50$ to $10\mytimes20$. EM@K and Recall@K are EM and Recall metrics measured by retrieving top $K/L$ sequences from $10\mytimes20$ candidates, where sequence scores are measured by Equation 1 and $L=2$ is the maximum number of hops in HotpotQA and StrategyQA.

\paragraph{Implementation Details}
All the experiments are conducted on $4\mytimes32$G V100 GPUs.
We initialize MoPo with a powerful pre-trained text embedder E5-v2-base \cite{wang2024textembeddingsweaklysupervisedcontrastive} as retrieval model. In addition, flan-t5-large \cite{chung2022scalinginstructionfinetunedlanguagemodels} is used as summary-generation model.

After training, we exploit the retrieval model $M_\theta$ to obtain document embeddings and index them with the Exact Inner Product (IndexFlatIP) in FAISS \cite{Johnson-2021-billion} for efficient search.
More implementation details are shown in Appendix \ref{train}.

\subsection{Baselines}

There are two groups of baselines for retrieval:

\paragraph{Models without posterior information} This group of baselines contains BeamDR \cite{zhao2021multistepreasoningunstructuredtext}, {MDR}$_{origin}$  \cite{xiong2021answeringcomplexopendomainquestions}, and several variants of MDR. Here, MDR is the variant of the original one where we exploit E5-v2-base as the hop dense retrieval instead of RoBERTa \cite{roberta} in the original MDR. 
{MDR$_{sum}$} and { {MDR}$_{zero}$} are the variants of MDR with two modifications: 1) The hop retrieval is E5-v2-base; 2) We exploit the summary model as in MoPo for query formalization instead of concatenating queries with retrieved documents as in MDR. Different from MDR and MDR$_{sum}$, the retrieval model of {MDR}$_{zero}$ are not finetuned on PostSumQA. The inclusion of {MDR}$_{zero}$ is for measuring the zero-shot performance when applying E5-v2-base as the hop retrieval in the MDR framework without further training. Note that once MoPo is trained, it is used for inference in the same way with MDR$_{sum}$. In other words, there is no additional inference cost associated with MoPo in comparison with MDR$_{sum}$. 

\paragraph{Models with Posterior Information} The baselines in this group includes {PR$_{fixed}$} and {PR$_{dyn}$}, which are MDR with two-stage training for posterior regularization \cite{chen-etal-2020-bridging}. {PR$_{fixed}$} \cite{chen-etal-2020-bridging} is trained with a fixed $\lambda\myeq0.3$, determined through grid search over the range $[0.1, 1.0]$ and a step size of $0.1$. For {PR$_{dyn}$}, we design a linear decay scheduler, analogous to learning rate scheduling, where $\lambda$ decreases from 0.3 to 0.1 over the course of training.

\subsection{Retrieval Results}

Table \ref{retrieval result} shows that MoPo outperforms all other baselines on both datasets. From the results, several findings can be obtained:
 {(1)} The fact that MDR is better than MDR$_{origin}$ shows that the adoption of the powerful (single-time) dense retrieval model, E5-v2-base, is essential for multi-hop retrieval. Notably, even MDR$_{zero}$ attains decent performance without training; 
 {(2)} MDR$_{sum}$ is significantly better than MDR, demonstrating that our query re-formalization is effective;
 {(3)} Although the dynamic adjustment of the proportion of PR term does enhance retrieval performance, both PR methods exhibit inferior performance compared to MDR$_{sum}$. PR$_{dyn}$ is better than PR$_{fixed}$ but worse than MoPo, showing that dynamic scheduling can help but not that much.
 {(4)} MoPo introduces a more effective training strategy for posterior knowledge distillation, resulting in superior performance compared to other baselines, particularly in EM metrics, which are critical for multi-hop QA systems.

Further evaluation of the held-out dataset, StrategyQA, provides some insightful observations:
{(1)} The superior performance of MDR$_{zero}$ over MDR, which is trained on HotpotQA, indicates that MDR exhibits limited generalization to StrategyQA. The possible reason is that MDR concatenates the query and the retrieved document for next-hop retrieval, making it prone to over-fitting to dataset-specific characteristics such as document length or irrelevant information. In contrast, MDR$_{sum}$ and MoPo, which use a summary model to rewrite the query, may avoid this issue.
{(2)} Despite being trained exclusively on PostSumQA, a derived set of HotpotQA training data, the summary model still facilitates retrieval performance, as being seen by the superior performance of MDR$_{sum}$ compared to MDR$_{zero}$;
{(3)} MoPo, trained on PostSumQA, also demonstrates its robustness when being tested on StrategyQA, significantly outperforming all other baselines in EM metrics.

 \begin{figure}[t]
     \centering
     \includegraphics[width=\linewidth]{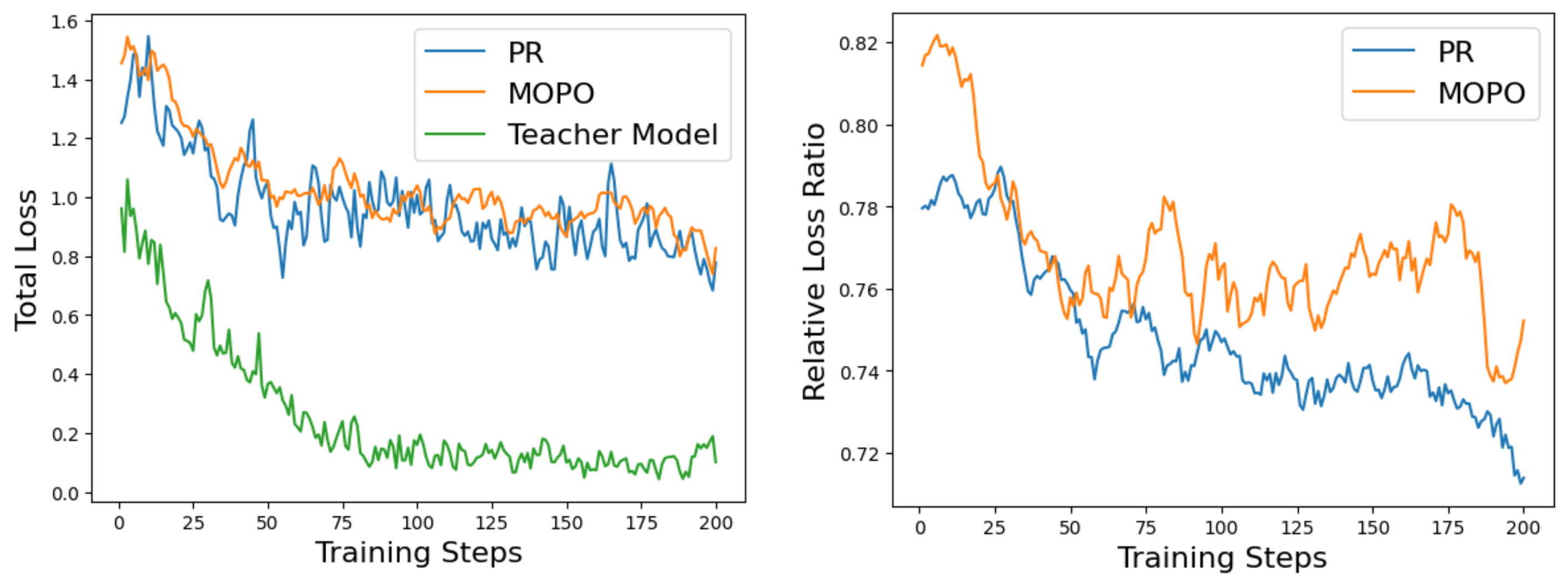}
 \caption{Total absolute and relative loss curve of PR and MoPo, $\lambda\myeq0.3$, where relative Loss Ratio = InfoNCE Loss / Total Loss. All curves are processed with the same smoothing factor.}
 \label{fig:loss_observation}
 \end{figure}
\subsection{Analysis of Training Loss Curve}\label{loss analysis}

Our experimental results show that {PR}$_{fixed}$ struggles to improve over {MDR}$_{sum}$ even with hyper-parameter tuning.
To further explore the reason, we draw the total and relative loss curves of {PR}$_{fixed}$ with $\lambda\myeq0.3$, the grid-searched value. Here, the relative loss measures the ratio of InfoNCE loss over the total loss (InfoNCE plus PR term). The lower the loss ratio is, the bigger the role of PR term is in optimization.

From the total loss plot in Figure \ref{fig:loss_observation}, we see that the posterior model (the teacher model) converges much faster compared to the student models. Specifically, the teacher model converges around $\app80$ steps while the student models (PR, MoPo) do not. From the relative loss plots, it is observable that  
the posterior regularization (PR) term always occupies a big ratio in the total loss. As training progresses, the proportion of PR terms increases faster than the decrease rate of total loss. 
By dynamically adjusting the weights of PR terms, we can mitigate this issue, resulting in a better results for {PR}$_{dyn}$ compared to PR$_{fixed}$ (see Table \ref{retrieval result}), yet the performance is still not satisfactory.
Figure \ref{fig:loss_observation} also shows that MoPo total loss is smoother compared to PR$_{fixed}$, and the higher relative loss curve indicates that MoPo can limit the domination of PR term during optimization. 

\subsection{Analysis on Hyper-parameter Sensitivity}
\begin{table}[t]
\centering
\scalebox{1}{
\begin{tabular}{p{1.3cm}cccc@{}}
\toprule
                & R@2   & R@20    & R@50      & R@100 \\ \midrule
\multicolumn{5}{c}{PR}  \\ \midrule
$\lambda=0.3$    & 94.49 & 95.07  & 95.88   & 96.55 \\
$\lambda=0.5$       & 93.85 & 94.39 & 95.13 & 95.75     \\
$\lambda=1.0$ & 93.93 & 94.59 & 95.29 & 95.85   \\
PG(\textperthousand)$~{\downarrow}$  & -6.773 & -7.153 & -7.822 & -8.286    \\
\midrule
\multicolumn{5}{c}{MoPo}  \\ \midrule
$\lambda=0.3$    & 94.76 & 95.42 & 96.17 & 96.85\\
$\lambda=0.5$       & 94.97 & 95.69 & 96.48 & 96.98   \\
$\lambda=1.0$ & 94.77 & 95.43 & 96.27 & 96.70    \\
PG(\textperthousand)$~{\downarrow}$         & \textbf{-2.211} & \textbf{-2.821}  & \textbf{-3.213}   & \textbf{-2.887}      \\
\bottomrule
\end{tabular}}
\caption{Retrieval performance on HotpotQA with different $\lambda$ values. PG means \textit{Performance Gap}, the ratio that deviates the most from the best performance among all performances} 
\label{lamda result}
\end{table}
\begin{table}[t]
\centering
\scalebox{1}{
\begin{tabular}{lcccc@{}}
\toprule
{Models}       & {R@2}   & {R@20}    & {R@50}      & {R@100} \\ \midrule
PR$_{fixed}$   & 94.49 & 95.07  & 95.88   & 96.55 \\
PR$_{dynamic}$  & 94.52 & 95.13  & 95.69   & 96.61 \\
\hline
$m=0$   & 94.53 & 95.27  & 95.92   & 96.62 \\

$m=0.5$    & 94.61 & \underline{95.36}  & 96.24   & \underline{96.68} \\
$m=0.9$       & 94.69 & 95.33 & \underline{96.25} & 96.67     \\
$m=0.99$ & \textbf{94.77} & \textbf{95.43} & \textbf{96.27} & \textbf{96.70}   \\
$m=1.0$     & \underline{94.73} & 95.33  & 96.21   & 96.67 \\
\bottomrule
\end{tabular}}
\caption{Retrieval performance of MoPo on HotpotQA with different momentum coefficient $m$. } 
\label{momentum result}
\end{table}
\label{mopo_analysis}
\paragraph{Less Sensitive to $\lambda$}
We study how the the performance of $PR_{fixed}$ and MoPo change with different values of the posterior regularization weight $\lambda$. The results in Table \ref{lamda result} show that MoPo has much lower performance gap compared to $PR_{fixed}$, verifying the robustness of MoPo in multi-hop dense retrieval. 

\paragraph{Momentum Update is More Effective}
We measure the impact of the parameter $m$ in Equation \ref{eq:mopo} on MoPo. 
Utilizing a grid search on a smaller validation set, we identified several representative values for testing as shown in Table \ref{momentum result}. 
Note that, $m\myeq0$ means the posterior model is the same as prior during every training step, and $m\myeq1$ means that the posterior model is kept to be the initial encoder during training.
As depicted in Table \ref{momentum result}, all models are better than PR$_{dynamic}$ on Recall@100.
Besides, $m\myeq0.99$ and $0$ performs the best and worst, respectively.
The best value of $m$ falls into the range between 0.9 and 1, indicating that MoPo prefers slower updating.

\section{Downstream Tasks}
\label{downstream task}
\begin{table}[t]
    \centering
    \begin{tabular}{lc}
        \toprule
        {Methods} & \multicolumn{1}{c}{{EM@2}} \\
        \midrule
        MDR (reranking) \cite{xiong2021answeringcomplexopendomainquestions} & 81.2 \\
        Beam Retrieval \cite{zhang2024endtoendbeamretrievalmultihop} & 82.2 \\
        Chain-of-Skills \cite{ma2023chainofskillsconfigurablemodelopendomain} & 88.9 \\
        MoPo (reranking)    & \textbf{89.4} \\
        \bottomrule
    \end{tabular}
    \caption{Fullwiki HotpotQA reranked retrieval results. The beam size here is $50\mytimes50$. Results of baseline methods come from their own paper. }
    \label{tab:rerank}
\end{table}

\begin{table*}[t]
    \centering
    \scalebox{0.85}{
    \begin{tabular}{lcccccccccccc}
        \toprule
        \multirow{2}{*}{Methods} & \multicolumn{6}{c}{Dev} & \multicolumn{6}{c}{Test} \\
        \cmidrule(lr){2-7} \cmidrule(lr){8-13}
        & \multicolumn{2}{c}{Ans} & \multicolumn{2}{c}{Sup} & \multicolumn{2}{c}{Joint} & \multicolumn{2}{c}{Ans} & \multicolumn{2}{c}{Sup} & \multicolumn{2}{c}{Joint} \\
        \cmidrule(lr){2-3} \cmidrule(lr){4-5} \cmidrule(lr){6-7} \cmidrule(lr){8-9} \cmidrule(lr){10-11} \cmidrule(lr){12-13}
        & EM & F1 & EM & F1 & EM & F1 & EM & F1 & EM & F1 & EM & F1\\
        \midrule
        MDR \cite{xiong2021answeringcomplexopendomainquestions} & 62.3 & 75.1 & 56.5 & 79.4 & 42.1 & 66.3 & 62.3 & 75.3 & 57.5 & 80.9 & 41.8 & 66.6 \\
        MDR(new) & 64.2 & 76.7 & 60.1 & 82.7 & 42.3 & 67.5 & - & - & - & - & - & - \\
        IRRR+ \cite{qi2021answeringopendomainquestionsvarying} & - & - & - & - & - & - & 66.3 & 79.9 & 57.2 & 82.6 & 43.1 & 69.8 \\
        HopRetriever-plus \cite{li2020hopretrieverretrievehopswikipedia} & 66.6 & 79.2 & 56.0 & 81.8 & 42.0 & 69.0 & 64.8 & 77.8 & 56.1 & 81.8 & 41.0 & 67.8 \\
        TPRR \cite{PathRanker} & 67.3 & 80.1 & 60.2 & 84.5 & 45.3 & 71.4 & 67.0 & 79.5 & 59.4 & 84.3 & 44.4 & 70.8 \\
        AISO \cite{zhu2021adaptiveinformationseekingopendomain} & 68.1 & 80.9 & 61.5 & 86.5 & 45.9 & 72.6 & 67.5 & 80.5 & 61.2 & 86.0 & 44.9 & 72.0 \\
        Chain-of-Skills \cite{ma2023chainofskillsconfigurablemodelopendomain} & 68.2 & 81.0 & 61.1 & 85.3 & 46.4 & 72.3 & 67.4 & 80.1 & 61.3 & 85.3 & \textbf{45.7} & 71.7 \\
        \midrule
        MoPo & \textbf{68.4} & \textbf{81.5} & \textbf{61.9} & \textbf{86.9} & \textbf{46.7} & \textbf{72.7} & \textbf{67.6} & \textbf{80.8} & \textbf{61.4} & \textbf{86.1} & \textbf{45.7} & \textbf{72.1} \\
        \bottomrule
    \end{tabular}
    }
    \caption{Downstream QA results on HotpotQA fullwiki set. MDR(new) means MDR reproduced with the same base models as MoPo. Results of other models come from their own paper.}
    \label{tab:QA}
\end{table*}

\begin{table}[t]
\centering

\scalebox{0.8}{
\begin{tabular}{lccc@{}}
\toprule
\textbf{Model}                & Overall   & Bridge    & Comp.  \\ \midrule
Self-Ask \cite{press-etal-2023-selfask}    & 49.4 & 45.3  & 68.6        \\
IRCoT \cite{trivedi-etal-2023-ircot}      & 56.2 & 53.4  & 69.6         \\
FLARE \cite{jiang-etal-2023-flare} & 56.1 & 54.2  & 54.4        \\
BeamAggR \cite{chu-etal-2024-beamaggr} & 62.9 & 60.5  & 74.2        \\
MoPo         & \textbf{64.3} & \textbf{61.2}  & \textbf{76.2}   \\
\bottomrule
\end{tabular}}
\caption{Downstream QA performance in F1 on HotpotQA subset of 100 samples \cite{trivedi-etal-2023-ircot}. Results of other models come from \cite{chu-etal-2024-beamaggr}.}
\label{ReasonQA}
\end{table}
\subsection{Reranking}
We utilize the pre-trained jina-reranker-v2-multilingual \cite{günther2023jina} as a reranker to test MoPo retrieval performance. 
After the retrieval stage, for each retrieved sequence $\{(d_1, d_2, \ldots, d_L)\}$, we use the reranker model to calculate the relevance score of each doc and calculate the sequence score as follows:
\begin{align}
    P(d_1, \ldots, d_L, q) & \propto \prod_{t}  P(d_t, q) \nonumber\\ 
    P(d_t, q) & = \operatorname{Reranker}([q, d_t]) \label{eq:reranker_socre}
\end{align}

Although this simple reranking assumes an independence assumption among documents in the sequence, its documents are retrieved in the sequencial order with DPR. In this experiment, we compare our results with two other competitive retrieval models that incorporate reranking techniques: {Beam Retrieval} \cite{zhang2024endtoendbeamretrievalmultihop}, {Chain-of-Skills (CoS)} \cite{ma2023chainofskillsconfigurablemodelopendomain}. Both Beam Retrieval and CoS exploit MDR as the retrieval framework and perform reranking at every hop. In contrast, we perform only ranking at the last hop. In addition, CoS incorporates multiple additional tasks, such as entity span prediction and linking, making the model much more complicated than MoPo+reranking. To be consistent with previous work, we set the beam size here at 50. In other words, for two hops in HotpotQA, the search space is $50\times50$. Experimental results in Table \ref{tab:rerank} demonstrate the advantages of MoPo over the baselines regardless of being more efficient compared to the stronger baselines (Beam Retrieval and CoS). It is noteworthy that, in comparison with CoS, MoPo achieves better performance with less computing requirement. Specifically, we need 4 V100 32G for training MoPo while CoS requires 16 V100 32G for training.

\subsection{Question Answering}
We evaluate MoPo when being used in a traditional retrieval-reranking-generation for QA task.  
Here, we utilize Flan-T5-large\cite{chung2022scalinginstructionfinetunedlanguagemodels} as the generation for answer and supporting sentence generation.

Table \ref{tab:QA} shows the results of MoPo and other baselines on the whole test set of HotpotQA with fullwiki setting, where MoPo establishes the state-of-the-art (SOTA) benchmark on both HotpotQA dev and test datasets. Furthermore, we observe a $1.2\%$ enhancement in performance when the base models of MDR are updated in MDR (new). Nevertheless, MoPo surpasses the revised MDR by a margin of $5.2\%$, which demonstrates the effectiveness of our methodology.

In addition, we compare MoPo-based pipeline with some other latest methods that exploit LLM for reasoning. We test on the same test set provided by IRCoT, which contains only 100 instances. Table\ref{ReasonQA} shows the results, where MoPo outperforms all baselines on HotpotQA set while maintaining a more efficient inference time.

\section{Conclusion}

This paper introduces Momentum Posterior Regularization (MoPo), the first attempt to distill posterior information into multi-hop dense retrieval. 
In MoPo, we define posterior information for each hop as a query-focused summary, and introduce a smooth and effective training strategy based on momentum-based moving average method. To facilitate MoPo training, we automatically construct PostSumQA from HotpotQA using a novel method, namely \textit{backward summary generation}. Our experimental results show the effectiveness of our method in exploiting posterior information, resulting in an improvement in retrieval performance. When being used in a traditional pipeline of retrieval-reranker-generation (Re2G), MoPo-based Re2G exhibits superior performance compared to strong baselines for two downstream tasks, reranking and QA.

\section*{Limitations}
The present study is subject to two principal limitations. 
First, although our training method is proven to be useful, the theory behind training controllability needs further exploration. We plan to further explore this in our future work.
Second, a comprehensive evaluation should include detailed analysis on the inference time of MoPo-based pipeline and other baselines in Reranking and QA tasks. However, the diversity of the baselines and the cost of such models prevent us to perform such analysis at the present. We aim to address these limitations in future research endeavors.

\section*{Ethics Statement}
MoPo aims to improve the performance of multi-hop dense retrieval and the training processing of posterior regularization.
We exclusively utilized existing datasets from previously published works. No new data collection was conducted for this study.
All experiments and query-focused summary constructions were performed strictly within the confines of these pre-existing datasets.
The nature of our generation process ensures that even in cases of inaccuracy, the outputs remain controllable and pose no potential harm. This is due to the constrained scope of the input data and the controlled nature of our experimental environment.
The current model operates solely in English, which inherently limits its practical applications in diverse, multilingual real-world scenarios.
\bibliography{custom}

\begin{thebibliography}{35}
\providecommand{\natexlab}[1]{#1}

\bibitem[{Chen et~al.(2020)Chen, Meng, Li, Chen, Xu, Xu, and Zhou}]{chen-etal-2020-bridging}
Xiuyi Chen, Fandong Meng, Peng Li, Feilong Chen, Shuang Xu, Bo~Xu, and Jie Zhou. 2020.
\newblock \href {https://doi.org/10.18653/v1/2020.emnlp-main.275} {Bridging the gap between prior and posterior knowledge selection for knowledge-grounded dialogue generation}.
\newblock In \emph{Proceedings of the 2020 Conference on Empirical Methods in Natural Language Processing (EMNLP)}, pages 3426--3437, Online. Association for Computational Linguistics.

\bibitem[{Chu et~al.(2024)Chu, Chen, Chen, Wang, Zhu, Du, Yu, Liu, and Qin}]{chu-etal-2024-beamaggr}
Zheng Chu, Jingchang Chen, Qianglong Chen, Haotian Wang, Kun Zhu, Xiyuan Du, Weijiang Yu, Ming Liu, and Bing Qin. 2024.
\newblock \href {https://aclanthology.org/2024.acl-long.67} {{B}eam{A}gg{R}: Beam aggregation reasoning over multi-source knowledge for multi-hop question answering}.
\newblock In \emph{Proceedings of the 62nd Annual Meeting of the Association for Computational Linguistics (Volume 1: Long Papers)}, pages 1229--1248, Bangkok, Thailand. Association for Computational Linguistics.

\bibitem[{Chung et~al.(2022)Chung, Hou, Longpre, Zoph, Tay, Fedus, Li, Wang, Dehghani, Brahma, Webson, Gu, Dai, Suzgun, Chen, Chowdhery, Castro-Ros, Pellat, Robinson, Valter, Narang, Mishra, Yu, Zhao, Huang, Dai, Yu, Petrov, Chi, Dean, Devlin, Roberts, Zhou, Le, and Wei}]{chung2022scalinginstructionfinetunedlanguagemodels}
Hyung~Won Chung, Le~Hou, Shayne Longpre, Barret Zoph, Yi~Tay, William Fedus, Yunxuan Li, Xuezhi Wang, Mostafa Dehghani, Siddhartha Brahma, Albert Webson, Shixiang~Shane Gu, Zhuyun Dai, Mirac Suzgun, Xinyun Chen, Aakanksha Chowdhery, Alex Castro-Ros, Marie Pellat, Kevin Robinson, Dasha Valter, Sharan Narang, Gaurav Mishra, Adams Yu, Vincent Zhao, Yanping Huang, Andrew Dai, Hongkun Yu, Slav Petrov, Ed~H. Chi, Jeff Dean, Jacob Devlin, Adam Roberts, Denny Zhou, Quoc~V. Le, and Jason Wei. 2022.
\newblock \href {https://arxiv.org/abs/2210.11416} {Scaling instruction-finetuned language models}.
\newblock \emph{Preprint}, arXiv:2210.11416.

\bibitem[{Demszky et~al.(2018)Demszky, Guu, and Liang}]{demszky2018transformingquestionansweringdatasets}
Dorottya Demszky, Kelvin Guu, and Percy Liang. 2018.
\newblock \href {https://arxiv.org/abs/1809.02922} {Transforming question answering datasets into natural language inference datasets}.
\newblock \emph{Preprint}, arXiv:1809.02922.

\bibitem[{Devlin et~al.(2019)Devlin, Chang, Lee, and Toutanova}]{devlin2019bertpretrainingdeepbidirectional}
Jacob Devlin, Ming-Wei Chang, Kenton Lee, and Kristina Toutanova. 2019.
\newblock \href {https://doi.org/10.18653/v1/N19-1423} {{BERT}: Pre-training of deep bidirectional transformers for language understanding}.
\newblock In \emph{Proceedings of the 2019 Conference of the North {A}merican Chapter of the Association for Computational Linguistics: Human Language Technologies, Volume 1 (Long and Short Papers)}, pages 4171--4186, Minneapolis, Minnesota. Association for Computational Linguistics.

\bibitem[{Feng et~al.(2020)Feng, Chen, Li, and Yin}]{Posterior-GAN}
Shaoxiong Feng, Hongshen Chen, Kan Li, and Dawei Yin. 2020.
\newblock Posterior-gan: Towards informative and coherent response generation with posterior generative adversarial network.
\newblock \emph{Proceedings of the AAAI Conference on Artificial Intelligence}, 34:7708--7715.

\bibitem[{Geva et~al.(2021)Geva, Khashabi, Segal, Khot, Roth, and Berant}]{geva2021strategyqa}
Mor Geva, Daniel Khashabi, Elad Segal, Tushar Khot, Dan Roth, and Jonathan Berant. 2021.
\newblock \href {https://doi.org/10.1162/tacl_a_00370} {Did aristotle use a laptop? a question answering benchmark with implicit reasoning strategies}.
\newblock \emph{Transactions of the Association for Computational Linguistics}, 9:346--361.

\bibitem[{Günther et~al.(2023)Günther, Ong, Mohr, Abdessalem, Abel, Akram, Guzman, Mastrapas, Sturua, Wang, Werk, Wang, and Xiao}]{günther2023jina}
Michael Günther, Jackmin Ong, Isabelle Mohr, Alaeddine Abdessalem, Tanguy Abel, Mohammad~Kalim Akram, Susana Guzman, Georgios Mastrapas, Saba Sturua, Bo~Wang, Maximilian Werk, Nan Wang, and Han Xiao. 2023.
\newblock \href {https://arxiv.org/abs/2310.19923} {Jina embeddings 2: 8192-token general-purpose text embeddings for long documents}.
\newblock \emph{Preprint}, arXiv:2310.19923.

\bibitem[{Hu et~al.(2022)Hu, Shen, Wallis, Allen-Zhu, Li, Wang, Wang, and Chen}]{hu2021loralowrankadaptationlarge}
Edward~J Hu, Yelong Shen, Phillip Wallis, Zeyuan Allen-Zhu, Yuanzhi Li, Shean Wang, Lu~Wang, and Weizhu Chen. 2022.
\newblock \href {https://openreview.net/forum?id=nZeVKeeFYf9} {Lo{RA}: Low-rank adaptation of large language models}.
\newblock In \emph{International Conference on Learning Representations}.

\bibitem[{Izacard et~al.(2022)Izacard, Caron, Hosseini, Riedel, Bojanowski, Joulin, and Grave}]{izacard2022unsuperviseddenseinformationretrieval}
Gautier Izacard, Mathilde Caron, Lucas Hosseini, Sebastian Riedel, Piotr Bojanowski, Armand Joulin, and Edouard Grave. 2022.
\newblock \href {https://arxiv.org/abs/2112.09118} {Unsupervised dense information retrieval with contrastive learning}.
\newblock \emph{Preprint}, arXiv:2112.09118.

\bibitem[{Jiang et~al.(2023)Jiang, Xu, Gao, Sun, Liu, Dwivedi-Yu, Yang, Callan, and Neubig}]{jiang-etal-2023-flare}
Zhengbao Jiang, Frank Xu, Luyu Gao, Zhiqing Sun, Qian Liu, Jane Dwivedi-Yu, Yiming Yang, Jamie Callan, and Graham Neubig. 2023.
\newblock \href {https://doi.org/10.18653/v1/2023.emnlp-main.495} {Active retrieval augmented generation}.
\newblock In \emph{Proceedings of the 2023 Conference on Empirical Methods in Natural Language Processing}, pages 7969--7992, Singapore. Association for Computational Linguistics.

\bibitem[{Johnson et~al.(2021)Johnson, Douze, and Jégou}]{Johnson-2021-billion}
Jeff Johnson, Matthijs Douze, and Hervé Jégou. 2021.
\newblock \href {https://doi.org/10.1109/TBDATA.2019.2921572} {Billion-scale similarity search with gpus}.
\newblock \emph{IEEE Transactions on Big Data}, 7(3):535--547.

\bibitem[{Karpukhin et~al.(2020)Karpukhin, Oguz, Min, Lewis, Wu, Edunov, Chen, and Yih}]{related_dpr_2020}
Vladimir Karpukhin, Barlas Oguz, Sewon Min, Patrick Lewis, Ledell Wu, Sergey Edunov, Danqi Chen, and Wen-tau Yih. 2020.
\newblock \href {https://doi.org/10.18653/v1/2020.emnlp-main.550} {Dense passage retrieval for open-domain question answering}.
\newblock In \emph{Proceedings of the 2020 Conference on Empirical Methods in Natural Language Processing (EMNLP)}, pages 6769--6781, Online. Association for Computational Linguistics.

\bibitem[{Kim et~al.(2020)Kim, Ahn, and Kim}]{kim2020sequentiallatentknowledgeselection}
Byeongchang Kim, Jaewoo Ahn, and Gunhee Kim. 2020.
\newblock \href {https://openreview.net/forum?id=Hke0K1HKwr} {{Sequential Latent Knowledge Selection for Knowledge-Grounded Dialogue}}.
\newblock In \emph{ICLR}.

\bibitem[{Kingma and Ba(2015)}]{adam_2015}
Diederik~P. Kingma and Jimmy Ba. 2015.
\newblock \href {http://arxiv.org/abs/1412.6980} {Adam: {A} method for stochastic optimization}.
\newblock In \emph{3rd International Conference on Learning Representations, {ICLR} 2015, San Diego, CA, USA, May 7-9, 2015, Conference Track Proceedings}.

\bibitem[{Li et~al.(2021)Li, Li, Shang, Jiang, Liu, Sun, Ji, and Liu}]{li2020hopretrieverretrievehopswikipedia}
Shaobo Li, Xiaoguang Li, Lifeng Shang, Xin Jiang, Qun Liu, Chengjie Sun, Zhenzhou Ji, and Bingquan Liu. 2021.
\newblock Hopretriever: Retrieve hops over wikipedia to answer complex questions.
\newblock In \emph{Proceedings of the AAAI conference on artificial intelligence}, volume~35, pages 13279--13287.

\bibitem[{Liu et~al.(2019)Liu, Ott, Goyal, Du, Joshi, Chen, Levy, Lewis, Zettlemoyer, and Stoyanov}]{roberta}
Yinhan Liu, Myle Ott, Naman Goyal, Jingfei Du, Mandar Joshi, Danqi Chen, Omer Levy, Mike Lewis, Luke Zettlemoyer, and Veselin Stoyanov. 2019.
\newblock \href {https://arxiv.org/abs/1907.11692} {Roberta: {A} robustly optimized {BERT} pretraining approach}.
\newblock \emph{CoRR}, abs/1907.11692.

\bibitem[{Loshchilov and Hutter(2019)}]{loshchilov2019decoupledweightdecayregularization}
Ilya Loshchilov and Frank Hutter. 2019.
\newblock \href {https://openreview.net/forum?id=Bkg6RiCqY7} {Decoupled weight decay regularization}.
\newblock In \emph{International Conference on Learning Representations}.

\bibitem[{Ma et~al.(2023)Ma, Cheng, Zhang, Liu, Nyberg, and Gao}]{ma2023chainofskillsconfigurablemodelopendomain}
Kaixin Ma, Hao Cheng, Yu~Zhang, Xiaodong Liu, Eric Nyberg, and Jianfeng Gao. 2023.
\newblock \href {https://doi.org/10.18653/v1/2023.acl-long.89} {Chain-of-skills: A configurable model for open-domain question answering}.
\newblock In \emph{Proceedings of the 61st Annual Meeting of the Association for Computational Linguistics (Volume 1: Long Papers)}, pages 1599--1618, Toronto, Canada. Association for Computational Linguistics.

\bibitem[{Muennighoff et~al.(2023)Muennighoff, Tazi, Magne, and Reimers}]{muennighoff2022mteb}
Niklas Muennighoff, Nouamane Tazi, Loic Magne, and Nils Reimers. 2023.
\newblock \href {https://doi.org/10.18653/v1/2023.eacl-main.148} {{MTEB}: Massive text embedding benchmark}.
\newblock In \emph{Proceedings of the 17th Conference of the European Chapter of the Association for Computational Linguistics}, pages 2014--2037, Dubrovnik, Croatia. Association for Computational Linguistics.

\bibitem[{Press et~al.(2023)Press, Zhang, Min, Schmidt, Smith, and Lewis}]{press-etal-2023-selfask}
Ofir Press, Muru Zhang, Sewon Min, Ludwig Schmidt, Noah Smith, and Mike Lewis. 2023.
\newblock \href {https://doi.org/10.18653/v1/2023.findings-emnlp.378} {Measuring and narrowing the compositionality gap in language models}.
\newblock In \emph{Findings of the Association for Computational Linguistics: EMNLP 2023}, pages 5687--5711, Singapore. Association for Computational Linguistics.

\bibitem[{Qi et~al.(2021)Qi, Lee, Sido, and Manning}]{qi2021answeringopendomainquestionsvarying}
Peng Qi, Haejun Lee, Tg~Sido, and Christopher Manning. 2021.
\newblock \href {https://doi.org/10.18653/v1/2021.emnlp-main.292} {Answering open-domain questions of varying reasoning steps from text}.
\newblock In \emph{Proceedings of the 2021 Conference on Empirical Methods in Natural Language Processing}, pages 3599--3614, Online and Punta Cana, Dominican Republic. Association for Computational Linguistics.

\bibitem[{Rajpurkar et~al.(2016)Rajpurkar, Zhang, Lopyrev, and Liang}]{rajpurkar-etal-2016-squad}
Pranav Rajpurkar, Jian Zhang, Konstantin Lopyrev, and Percy Liang. 2016.
\newblock \href {https://doi.org/10.18653/v1/D16-1264} {{SQ}u{AD}: 100,000+ questions for machine comprehension of text}.
\newblock In \emph{Proceedings of the 2016 Conference on Empirical Methods in Natural Language Processing}, pages 2383--2392, Austin, Texas. Association for Computational Linguistics.

\bibitem[{Thakur et~al.(2021)Thakur, Reimers, R{\"u}ckl{\'e}, Srivastava, and Gurevych}]{thakur2021beirheterogenousbenchmarkzeroshot}
Nandan Thakur, Nils Reimers, Andreas R{\"u}ckl{\'e}, Abhishek Srivastava, and Iryna Gurevych. 2021.
\newblock \href {https://openreview.net/forum?id=wCu6T5xFjeJ} {{BEIR}: A heterogeneous benchmark for zero-shot evaluation of information retrieval models}.
\newblock In \emph{Thirty-fifth Conference on Neural Information Processing Systems Datasets and Benchmarks Track (Round 2)}.

\bibitem[{Trivedi et~al.(2023)Trivedi, Balasubramanian, Khot, and Sabharwal}]{trivedi-etal-2023-ircot}
Harsh Trivedi, Niranjan Balasubramanian, Tushar Khot, and Ashish Sabharwal. 2023.
\newblock \href {https://doi.org/10.18653/v1/2023.acl-long.557} {Interleaving retrieval with chain-of-thought reasoning for knowledge-intensive multi-step questions}.
\newblock In \emph{Proceedings of the 61st Annual Meeting of the Association for Computational Linguistics (Volume 1: Long Papers)}, pages 10014--10037, Toronto, Canada. Association for Computational Linguistics.

\bibitem[{van~den Oord et~al.(2019)van~den Oord, Li, and Vinyals}]{oord2019representationlearningcontrastivepredictive}
Aaron van~den Oord, Yazhe Li, and Oriol Vinyals. 2019.
\newblock \href {https://arxiv.org/abs/1807.03748} {Representation learning with contrastive predictive coding}.
\newblock \emph{Preprint}, arXiv:1807.03748.

\bibitem[{Wang et~al.(2024)Wang, Yang, Huang, Jiao, Yang, Jiang, Majumder, and Wei}]{wang2024textembeddingsweaklysupervisedcontrastive}
Liang Wang, Nan Yang, Xiaolong Huang, Binxing Jiao, Linjun Yang, Daxin Jiang, Rangan Majumder, and Furu Wei. 2024.
\newblock \href {https://arxiv.org/abs/2212.03533} {Text embeddings by weakly-supervised contrastive pre-training}.
\newblock \emph{Preprint}, arXiv:2212.03533.

\bibitem[{Wei et~al.(2022)Wei, Wang, Schuurmans, Bosma, Ichter, Xia, Chi, Le, and Zhou}]{cot_wei_nips}
Jason Wei, Xuezhi Wang, Dale Schuurmans, Maarten Bosma, Brian Ichter, Fei Xia, Ed~H. Chi, Quoc~V. Le, and Denny Zhou. 2022.
\newblock Chain-of-thought prompting elicits reasoning in large language models.
\newblock In \emph{Advances in Neural Information Processing Systems 35: Annual Conference on Neural Information Processing Systems 2022, NeurIPS 2022, New Orleans, LA, USA, November 28 - December 9, 2022}.

\bibitem[{Xia et~al.(2023)Xia, Gou, Yu, Yu, Huang, Li, and Cam-Tu}]{xia2023improvingquestiongenerationmultilevel}
Zehua Xia, Qi~Gou, Bowen Yu, Haiyang Yu, Fei Huang, Yongbin Li, and Nguyen Cam-Tu. 2023.
\newblock \href {https://doi.org/10.18653/v1/2023.findings-emnlp.57} {Improving question generation with multi-level content planning}.
\newblock In \emph{Findings of the Association for Computational Linguistics: EMNLP 2023}, pages 800--814, Singapore. Association for Computational Linguistics.

\bibitem[{Xiong et~al.(2021)Xiong, Li, Iyer, Du, Lewis, Wang, Mehdad, Yih, Riedel, Kiela, and O{\u{g}}uz}]{xiong2021answeringcomplexopendomainquestions}
Wenhan Xiong, Xiang~Lorraine Li, Srinivasan Iyer, Jingfei Du, Patrick Lewis, William~Yang Wang, Yashar Mehdad, Wen-tau Yih, Sebastian Riedel, Douwe Kiela, and Barlas O{\u{g}}uz. 2021.
\newblock Answering complex open-domain questions with multi-hop dense retrieval.
\newblock \emph{International Conference on Learning Representations}.

\bibitem[{Yang et~al.(2018)Yang, Qi, Zhang, Bengio, Cohen, Salakhutdinov, and Manning}]{yang2018hotpotqa}
Zhilin Yang, Peng Qi, Saizheng Zhang, Yoshua Bengio, William Cohen, Ruslan Salakhutdinov, and Christopher~D. Manning. 2018.
\newblock \href {https://doi.org/10.18653/v1/D18-1259} {{H}otpot{QA}: A dataset for diverse, explainable multi-hop question answering}.
\newblock In \emph{Proceedings of the 2018 Conference on Empirical Methods in Natural Language Processing}, pages 2369--2380, Brussels, Belgium. Association for Computational Linguistics.

\bibitem[{Zhang et~al.(2024)Zhang, Zhang, Zhang, Yong, and Huang}]{zhang2024endtoendbeamretrievalmultihop}
Jiahao Zhang, Haiyang Zhang, Dongmei Zhang, Liu Yong, and Shen Huang. 2024.
\newblock \href {https://doi.org/10.18653/v1/2024.naacl-long.96} {End-to-end beam retrieval for multi-hop question answering}.
\newblock In \emph{Proceedings of the 2024 Conference of the North American Chapter of the Association for Computational Linguistics: Human Language Technologies (Volume 1: Long Papers)}, pages 1718--1731, Mexico City, Mexico. Association for Computational Linguistics.

\bibitem[{Zhang et~al.(2021)Zhang, Zhan, Hu, Fu, Luo, Jiang, Jia, Yu, Dou, Cao, and Chen}]{PathRanker}
Xinyu Zhang, Ke~Zhan, Enrui Hu, Chengzhen Fu, Lan Luo, Hao Jiang, Yantao Jia, Fan Yu, Zhicheng Dou, Zhao Cao, and Lei Chen. 2021.
\newblock \href {https://doi.org/10.1145/3404835.3462942} {Answer complex questions: Path ranker is all you need}.
\newblock In \emph{Proceedings of the 44th International ACM SIGIR Conference on Research and Development in Information Retrieval}, SIGIR '21, page 449–458, New York, NY, USA. Association for Computing Machinery.

\bibitem[{Zhao et~al.(2021)Zhao, Xiong, Boyd-Graber, and Daum{\'e}~III}]{zhao2021multistepreasoningunstructuredtext}
Chen Zhao, Chenyan Xiong, Jordan Boyd-Graber, and Hal Daum{\'e}~III. 2021.
\newblock \href {https://doi.org/10.18653/v1/2021.naacl-main.368} {Multi-step reasoning over unstructured text with beam dense retrieval}.
\newblock In \emph{Proceedings of the 2021 Conference of the North American Chapter of the Association for Computational Linguistics: Human Language Technologies}, pages 4635--4641, Online. Association for Computational Linguistics.

\bibitem[{Zhu et~al.(2021)Zhu, Pang, Lan, Shen, and Cheng}]{zhu2021adaptiveinformationseekingopendomain}
Yunchang Zhu, Liang Pang, Yanyan Lan, Huawei Shen, and Xueqi Cheng. 2021.
\newblock \href {https://doi.org/10.18653/v1/2021.emnlp-main.293} {Adaptive information seeking for open-domain question answering}.
\newblock In \emph{Proceedings of the 2021 Conference on Empirical Methods in Natural Language Processing}, pages 3615--3626, Online and Punta Cana, Dominican Republic. Association for Computational Linguistics.

\end{thebibliography}

\appendix
\setcounter{figure}{0}
\renewcommand{\thefigure}{A\arabic{figure}}
\setcounter{table}{0}
\renewcommand{\thetable}{A\arabic{table}}

\section{Methods and Analyses of Data Synthesis}
\label{sec:Data Synthesis}

\subsection{Detailed pipeline of data synthesis}
\paragraph{Rule-based method: QA2D} The second-hop summaries of all bridge questions are generated by the rule-based method QA2D\cite{demszky2018transformingquestionansweringdatasets}. The original intention of this method is to create a large-scale semi-supervised NLI (Natural Language Inference) dataset. Compared with the single NLI dataset, there are more publicly available manually annotated QA datasets. QA2D can convert question-answer pairs into corresponding declarative forms. Although the original QA2D paper only conducted experiments on SQuAD\cite{rajpurkar-etal-2016-squad}, Xia \cite{xia2023improvingquestiongenerationmultilevel}also verified the effectiveness of this method on the HotpotQA dataset. The inherent disadvantage of rule-based methods is poor generalization. The answers to comparison-type question-answer pairs in the HotpotQA dataset are mainly "yes" and "no", while SQuAD is mainly factoid question answering, and the answers are extractive. QA2D developed on SQuAD cannot cover all examples in HotpotQA, especially comparison-type question answering pairs. Therefore, this paper only uses this method to generate the second-hop summary of the bridging problem. If we encounter a sample that QA2D cannot calculate, we will directly discard it.
\paragraph{LLM-based method} We use GPT-3.5-Turbo to complete the first-hop and second-hop summary generation of all comparison type samples, and the first-hop summary generation task of all bridge type samples. We write the prompt words for summary generation according to the following points, taking the prompt words for the first-hop summary generation of bridge type samples as an example:
\begin{enumerate}
    \item Complex task decomposition: The query-centric summary generation task with multiple thinking steps is decomposed into multiple subtasks, namely reasoning step generation and query-centric summary generation.
    \item The requirements should be clear. Use Markdown syntax to write (1) a detailed explanation of each item entered; (2) clearly state each step required to complete the task;
    \item Two examples are provided: the examples are input in order: \texttt{Context}, \texttt{Statement}, \texttt{Reasoning steps} and \texttt{Output}. The context only uses the title and crowd-sourced supporting fact sentences instead of the entire supporting paragraph to provide more effective information.
    \item Write the steps of the Chain of Thought into the example. And the output should include the Reasoning Steps.
    \item If an input example is too difficult for the LLM, skip that example.
\end{enumerate}
We provide a detailed example in Appendix \ref{sec:Data Synthesis Example}.

\paragraph{Semi-supervised data generation method for comparing type samples} Considering that comparison type problems are relatively difficult problems, we add a specific interpretation of comparison type problems to the prompt words for the first-hop summary generation of bridge type samples, and modify the execution steps and requirements of specific tasks. Considering that multi-task text generation, that is, it is difficult to generate the first-hop and second-hop summaries at the same time, and to avoid the mutual influence of the generation results, we perform two tasks separately. The input for the first-hop summary generation is the input question and a supporting document, and the input for the second-hop summary generation is the question, the answer, and two supporting documents. The reasoning step requirements in the prompt words of the two tasks are shown in Figure \ref{fig:gpt_cot_comparison}. In general, the specific interpretation of comparison type problems has been given, and the overall idea of the two generation tasks is to find common content first and then generate summaries.

\subsection{Data synthesis example}
\label{sec:Data Synthesis Example}
\begin{figure}[t]
    \centering
    \lstset{
 columns=fixed,       
 frame=none,                                          
 backgroundcolor=\color[RGB]{245,245,244},            
 keywordstyle=\color[RGB]{40,40,255},                 
 numberstyle=\footnotesize\color{darkgray},           
 commentstyle=\it\color[RGB]{0,96,96},                
 stringstyle=\rmfamily\slshape\color[RGB]{128,0,0},   
 showstringspaces=false,                              
 language=python,                                        
}
    \begin{lstlisting}[language=python, breaklines=true]
"""You're a content writer, and your task is to re-write the ###Statement by removing the contents that ###Context does not mentions.

The input contains:
###Context: A title and its corresponding sentences.
###Statement: A statement but contains the information that is not appears in the , ###Context and maybe has some grammar errors. 

You can finish this as follow steps:
1. Find the common contents that both ###Context and ###Statement contain.
2. Remove the noise information that is not appears in ###Context.
3. Fix the grammar error, and keep content consistency.
"""
    \end{lstlisting}
    \caption{Main content of the prompt words generated by the first hop summary of the bridge type sample}
    \label{fig:gpt_bridge_prompts}
\end{figure}
\begin{figure}[t]
    \centering
    \lstset{
 columns=fixed,       
 frame=none,                                          
 backgroundcolor=\color[RGB]{245,245,244},            
 keywordstyle=\color[RGB]{40,40,255},                 
 numberstyle=\footnotesize\color{darkgray},           
 commentstyle=\it\color[RGB]{0,96,96},                
 stringstyle=\rmfamily\slshape\color[RGB]{128,0,0},   
 showstringspaces=false,                              
 language=python,                                        
}
    \begin{lstlisting}[language=python, breaklines=true]
""" 1. Common contents: Both the context and statement mention Jun Li and his association with Harvard University.
2. Removing irrelevant information: The statement mentions the Fields Medal received in 1982, but this information is not present in the context. Therefore, it is removed from the output.
3. Grammar and content consistency: The output sentence is rephrased to maintain grammatical correctness and content consistency. It states that Shing-Tung Yau was the PhD supervisor of Jun Li at Harvard University, which is the relevant information mentioned in the context."""
    \end{lstlisting}
    
    \caption{Example of reasoning steps for first-hop summary generation in bridge-type question answering}
    \label{fig:gpt_cot_bridge}
\end{figure}
\paragraph{LLM-based method example} Figure \ref{fig:gpt_bridge_prompts} shows the prompt words for generating the first-hop summary of the bridge type. For the sake of convenience, the examples and input forms in the prompt words are ignored. The reasoning chain of the example is completed using GPT and manually corrected. Taking the content in Figure \ref{fig:fig1} as an example, the reasoning process for generating the first-hop summary is shown in Figure \ref{fig:gpt_cot_bridge}, and its reasoning steps are strictly carried out in accordance with the requirements in Figure \ref{fig:gpt_bridge_prompts}. Intuitively, this can ensure the consistency of the generated content. The specific explanations of Figure \ref{fig:gpt_bridge_prompts} and Figure \ref{fig:gpt_cot_bridge} are as follows:
\begin{itemize}
    \item First, find the common content between the context and the complete answer, mainly "\textit{Jun Li}" and "\textit{Harvard University}";
    \item Second, find the content mentioned in the complete answer but not covered by the context, which is "\textit{Fields Medal received in 1982}", and remove it from the complete answer;
    \item Finally, considering that the complete answer is based on the output of the rule model, there may be minor errors such as grammar, which can be corrected using LLM.
\end{itemize}
The method shown in Figure \ref{fig:gpt_bridge_prompts} does not directly use the context and the given question to directly generate the first-hop summary, but instead inputs the complete answer constructed based on the QA2D method, which is equivalent to removing redundant content in the complete answer based on the context content, that is, content not included in the context.
\begin{figure}[t]
    \centering
    \lstset{
 columns=fixed,       
 frame=none,                                          
 backgroundcolor=\color[RGB]{245,245,244},            
 keywordstyle=\color[RGB]{40,40,255},                 
 numberstyle=\footnotesize\color{darkgray},           
 commentstyle=\it\color[RGB]{0,96,96},                
 stringstyle=\rmfamily\slshape\color[RGB]{128,0,0},   
 showstringspaces=false,                              
 language=python,                                        
}
    \begin{lstlisting}[language=python, breaklines=true]
First_hop = """
1. Identify what the ###Question asks, and find the relevant information in the ###Context.
2. Identify the common contents that the ###Context contains and the ###Question asks.
3. Ingore or move the information that ###Question asks but ###Context does not mention.
4. According to the common contents, write a question-oriented statement of the ###Context."""

Second_hop = """
1. First check the ###Answer is yes/no or some specific contents.
2. Identify the relevant content to the ###Question ask in ###Context1.
3. Identify the relevant content to the ###Question ask in ###Context2.
4. Combine the ###Answer and ###Question according to the relevant content"""
    \end{lstlisting}
    \caption{Main content of inference hint words for summary generation of comparative question-answer pairs}
    \label{fig:gpt_cot_comparison}
\end{figure}
\subsection{Quality analysis of LLM-filtering data}
\label{sec: Mistake analysis}
\begin{table}[t]
    \centering
    \small

    \begin{tabular}{p{2cm}p{4.5cm}}
\toprule
\textbf{Key} & \multicolumn{1}{c}{\textbf{Content}} \\ \midrule
Question          & Who was an American former professional baseball left fielder, Andrea   Martin or Chuck Essegian?  \\ \midrule
Context        & Andrea Martin. Andrea Louise Martin (born January 15, 1947) is an   American actress, singer, author and comedian, best known for her work in the   television series ''SCTV''. \\ \hline
\multirow{4}{*}{Decline Reason} & 
(1) Identify what the Question asks: The Question asks for a comparison of nationality between two individuals, Andrea Martin and Chuck Essegian.\\ 
 & (2) Identify the common content between the Context and the Question: Andrea Martin.\\
 & (3) Remove the information that the Question asks but the Context does not mention: Chuck Essegian. \\
 & (4) No common content found between the Context and the Question. \\\hline
 2nd Hop summary & Charles Abraham 'Chuck' Essegian was an American former professional baseball left fielder, while Andrea Martin is an American actress, singer, author, and comedian.\\
\bottomrule
\end{tabular}
\caption{Example of first-hop summary generation failure in LLM filter comparison type}
\label{tab:gpt_cmp_bad_case}
\end{table}
\begin{table}[t]
    \centering
\begin{tabular}{lcc}
\toprule
\textbf{Stage}    & \textbf{Bridge}           & \textbf{Comparison}              \\ \midrule
Original    & 20165(80.66\%) & 4835(19.34\%)  \\
Rule-filtering & 19398(80.11\%) & 4815(19.87\%)  \\
LLM-filtering & 19120(84.24\%) & 3576(15.76\%)  \\ \bottomrule
\end{tabular}
\caption{Distribution of two types of questions in the semi-supervised data production process}
\label{data_type}
\end{table}
In Table \ref{data_type}, rule filtering primarily refers to GPT’s string outputs unaccounted for by QA2D rules or unparseable in \texttt{JSON} format; LLM filtering denotes tasks that GPT deems unanswerable based on input instructions. The sample type distribution reveals that rules filtered approximately $3\%$ of the data, while LLM filtered around $9\%$, predominantly comparison type data. This could stem from the increased complexity of generating comparison type summaries compared to bridge type, with LLM potentially struggling to comprehend or extract the compared entities. 

Considering that the proportion of samples filtered out by LLM in Table \ref{data_type} is relatively high, we further analyze the proportion of first-hop and second-hop summary generation failures in failed samples. We found that LLM filtered $1210$ first-hop summaries and $59$ second-hop summaries. The loss of $59$ second-hop samples is normal data iteration loss, so we manually screened the reasons for the failure of the first-hop samples. Table \ref{tab:gpt_cmp_bad_case} shows an example of a comparison type where the first-hop summary generation failed. GPT made two misjudgments. The first was that the question was about nationality, and the second was that the given context and the question had no common content. The possible reason is that the input question specifies the specific content of the occupation, while the input context is only related to nationality, which is too difficult for GPT. When generating the second summary, both target paragraphs are used as input, so it is easier to generate the second-hop summary.

\begin{figure}
\centering
    \begin{subfigure}{.49\linewidth}
    \centering
    \includegraphics[width=\linewidth]{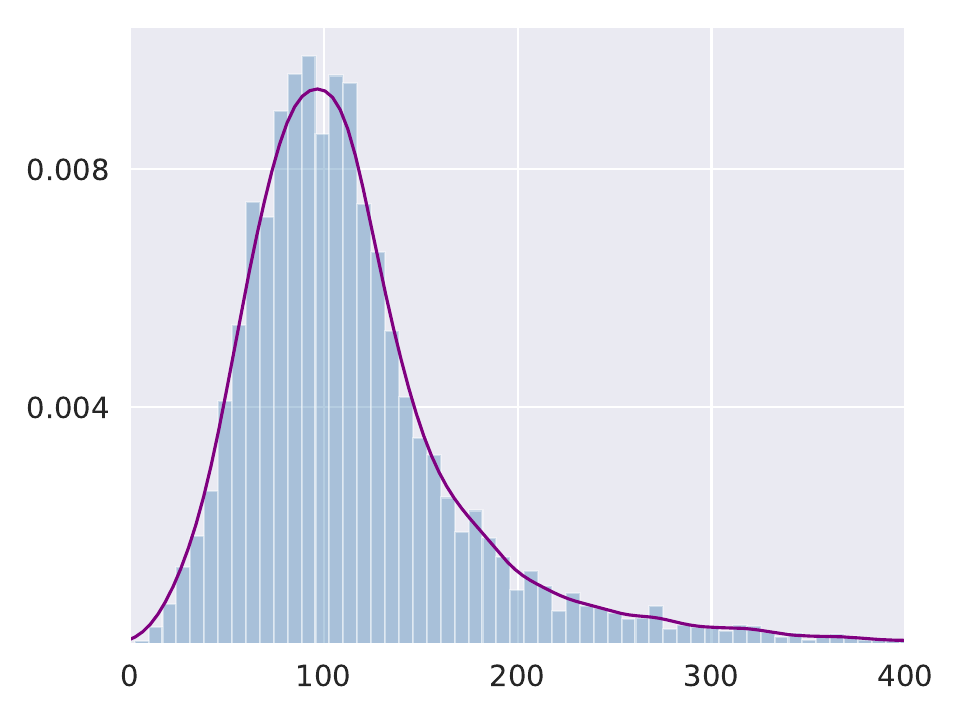}
    \caption{Ratio of first-hop summary to question length}
    \label{fig:chapter4_gpt_analysis_a}
    \end{subfigure}
    \begin{subfigure}{.49\linewidth}
    \centering
    \includegraphics[width=\linewidth]{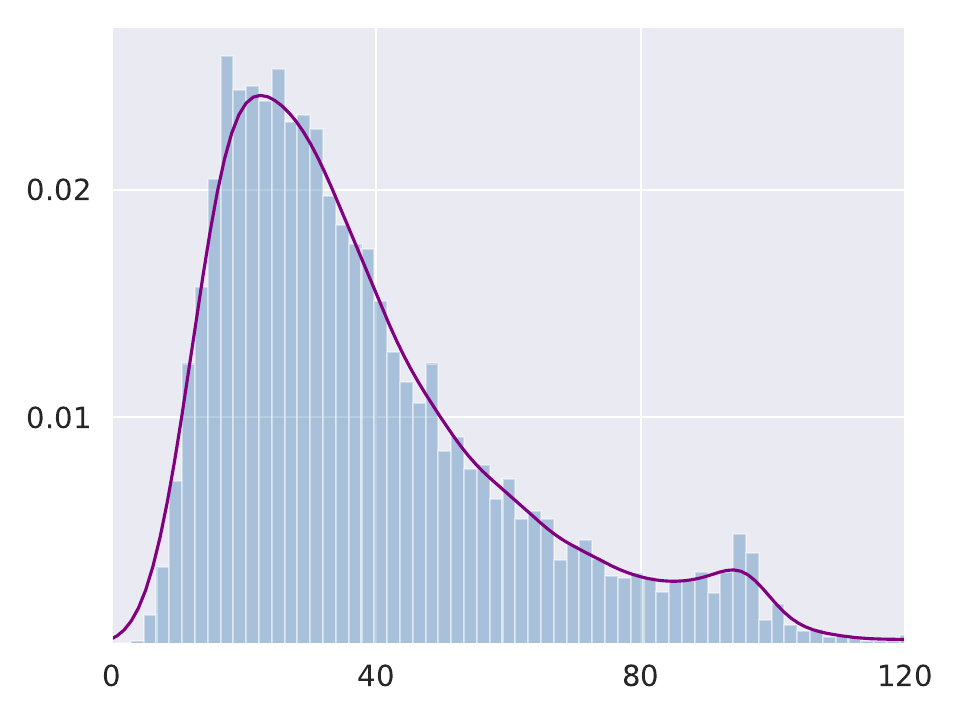}
    \caption{Ratio of first-hop summary to supporting facts}
    \label{fig:chapter4_gpt_analysis_b}
    \end{subfigure}
\caption{Relative length distribution of first-hop summaries generated using only GPT}
\label{fig:chapter4_gpt_analysis}
\end{figure}

\begin{figure}
\centering
    \begin{subfigure}{.49\linewidth}
    \centering
    \includegraphics[width=\linewidth]{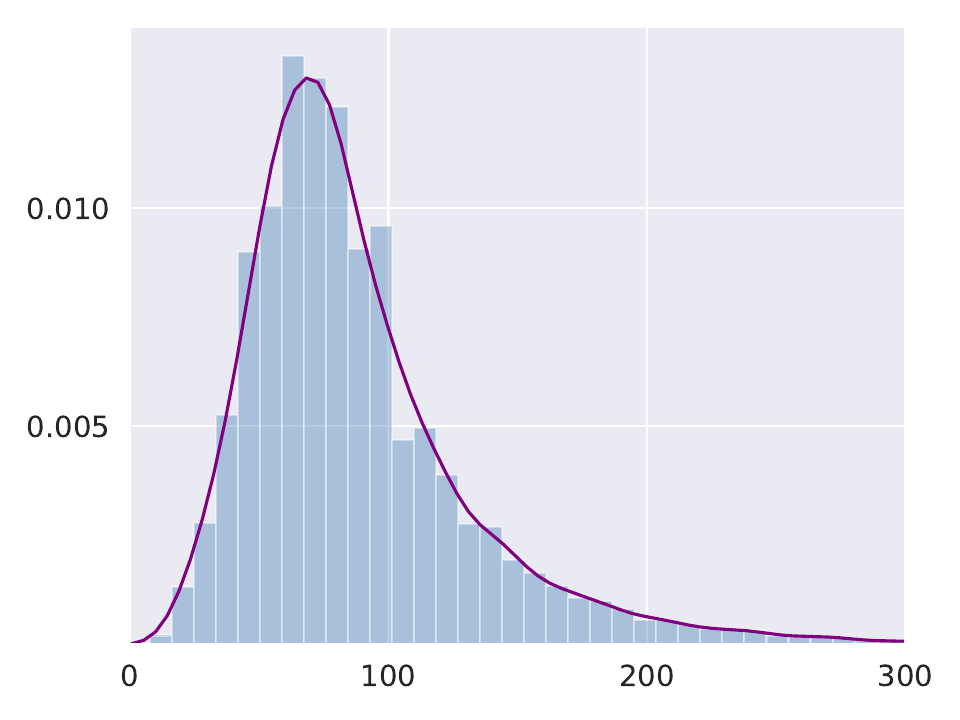}
    \caption{Ratio of first-hop summary to question length}
    \label{fig:chapter4_qa2d_analysis_a}
    \end{subfigure}
    \begin{subfigure}{.49\linewidth}
    \centering
    \includegraphics[width=\linewidth]{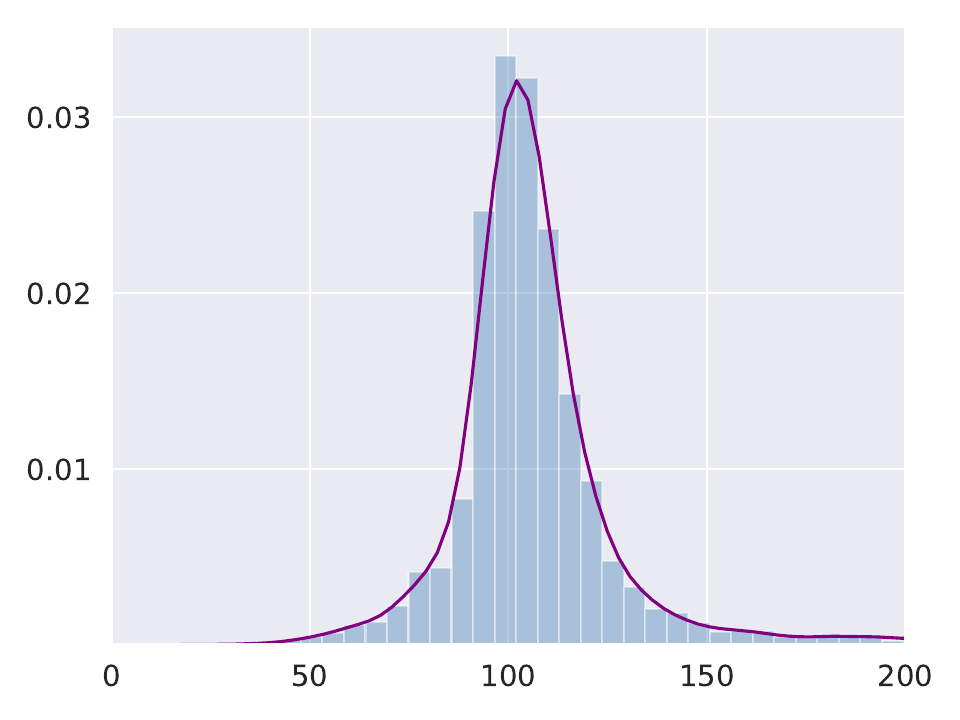}
    \caption{Ratio of the second hop summary to question length}
    \label{fig:chapter4_qa2d_analysis_b}
    \end{subfigure}
\caption{Relative length distribution of first- and second-hop summaries and questions in the semi-supervised dataset}
\label{fig:chapter4_qa2d_analysis}
\end{figure}

In addition, we also explored the effect of directly using GPT-3.5-Turbo to generate first-hop summaries. The first step of the input instruction is to summarize the common content, and the second step is to remove the content not mentioned in the input context.The results in Table \ref{dataset_eval} show that this is worse than the indirect method used in this paper: stripping the full answer into a first-hop summary.Manual evaluation found that the first-hop summary output by this method still carries content that is not mentioned in the input context but included in the question.
Figure \ref{fig:chapter4_gpt_analysis} shows $9559$ statistics of first-hop summary lengths.
As can be seen from Figure \ref{fig:chapter4_gpt_analysis_b}, the first-hop summary generated by GPT is shorter than the input long context in terms of text length, which is in line with the summary task scenario.
However, as can be seen from \ref{fig:chapter4_gpt_analysis_a}, the text length of the first-hop summary is almost the same as the input question. Considering that the input is a multi-hop question involving two related paragraphs, the statistical result is consistent with the result of manual evaluation.
In Figure\ref{fig:chapter4_qa2d_analysis}, the relative length distribution results of $22696$ semi-supervised data are shown.
Figure\ref{fig:chapter4_qa2d_analysis_a} shows that the relative length of the first-hop summary produced by the final method is concentrated around $0.8$ to $0.9$. Figure\ref{fig:chapter4_qa2d_analysis_b} shows that the relative length of the second-hop summary is closer to a normal distribution with a smaller variance, further verifying the effectiveness of our semi-supervised dataset production method.

\begin{table}[t]
    \centering
    
    \begin{tabular}{lcc}
\toprule
    & \textbf{Correctness}           & \textbf{Coverage}              \\ \midrule
PostSumQA    & 9.36 & 9.03  \\
GPT-generated & 9.23 & 8.52  \\
 \bottomrule
\end{tabular}
\caption{Human review between summaries in PostSumQA and GPT-generated data}
\label{dataset_eval}
\end{table}

\setcounter{figure}{0}
\renewcommand{\thefigure}{B\arabic{figure}}
\setcounter{table}{0}
\renewcommand{\thetable}{B\arabic{table}}

\section{Training Details}
\label{train}
\subsection{Implementation Details}
\paragraph{Base Model} We use E5-base\cite{wang2024textembeddingsweaklysupervisedcontrastive}, Jina-Reranker-v2\cite{günther2023jina} and Flan-T5-large\cite{chung2022scalinginstructionfinetunedlanguagemodels} as the base models for retrieval tasks, reranking tasks and query-centric summary generation tasks respectively. The network structure of E5-base is the same as BERT, with 110M parameters. Its pre-training data does not include HotpotQA, and its performance has achieved excellent results on MTEB\cite{muennighoff2022mteb} and BEIR\cite{thakur2021beirheterogenousbenchmarkzeroshot}. The Jina-Reranker-v2 is a transformer-based model that has been fine-tuned for text reranking task, which has demonstrated competitiveness across a series of benchmarks targeting for text retrieval against other reranker models. Flan-T5-large is a generative model that is fine-tuned based on T5-large and has excellent performance in various natural language processing tasks. We use a simple instruction template and removes the samples and specific explanations of the tasks in the previous data augmentation instructions.

\begin{table}[t]
    \centering
    \label{Training detail}
    \begin{subtable}[t]{\linewidth}
        \centering
        \begin{tabular}{ll}
        \hline
        Learning Rate & 2e-5 \\
        Data batch size & 128 \\
        Paragraph cutoff size & 350 \\
        Query cutoff size & 350 \\
        Learning rate warm-up rate & 0.1 \\
        Gradient clipping & 2.0 \\
        Training steps & 200 \\
        Weight Decay & 0.1 \\
        \hline
        \end{tabular}
        \caption{Retrieval model training hyperparameters}
        \label{Tab3a}
    \end{subtable}
    
    \vspace{0.3cm} 
    
    \begin{subtable}[t]{\linewidth}
        \centering
        \begin{tabular}{ll}
        \hline
        Learning Rate & 2e-5 \\
        Data batch size & 32 \\
        Gradient accumulation times & 4 \\
        Number of training rounds & 1 \\
        Learning rate warm-up rate & 0.1 \\
        Enter the cutoff length & 400 \\
        Maximum output length & 60 \\
        Whether to sample & No \\
        \hline
        \end{tabular}
        \caption{Query-centric summary generation model training hyperparameters}
        \label{Tab3b}
    \end{subtable}
    \caption{Training hyperparameter settings}
\end{table}

\paragraph{Parameter Setting} In the retrieval task, the baseline model MDR, the retrieval model enhanced by summary enhancement $m_{\theta}$, the teacher model $m_{\phi}$, the basic posterior regularization model (hereinafter referred to as PR) and MoPo are all trained for 200 steps. The momentum factor $m$ is set to 0.99. Other hyperparameters are shown in Table\ref{Tab3a}. In the query-centric summary generation task, two sample examples are used, which is consistent with the instruction format when semi-supervised data is produced. All linear layers in the attention mechanism are enhanced using LoRA\cite{hu2021loralowrankadaptationlarge} and only trained once. The training hyperparameters are shown in Table\ref{Tab3b}. All model optimizers are AdamW\cite{loshchilov2019decoupledweightdecayregularization}.
\subsection{Details of Evaluation Metrics}
\label{metric_detail}
\textbf{EM} measures whether the predicted answer exactly matches the ground truth answer. For a single query, if the predicted answer is identical to the ground truth (ignoring case and punctuation), the score is 1; otherwise, it is 0. The final EM score is the average of exact matches across all queries:  
      \[
      EM = \frac{\sum_{i=1}^{N} \mathbbm{1}(\text{Predict}_i = \text{Answer}_i)}{N}
      \]
      where \( N \) is the total number of queries.

\textbf{Recall} measures how many relevant answers from ground truth are retrieved by the model. For Top-K retrieval, Recall@K is calculated as:
      \[
      R@K = \frac{\sum_{i=1}^{N} \mathbbm{1}(\text{Answer}_i \in \text{Top-K Preds}_i)}{N}
      \]
      where \( N \) is the total number of queries, and Top-K Predictions are the top K returned answers.

\setcounter{figure}{0}
\renewcommand{\thefigure}{C\arabic{figure}}
\setcounter{table}{0}
\renewcommand{\thetable}{C\arabic{table}}

\section{Additional Experiments and Details}
\subsection{Ablation Study}
\begin{table}[t]
\centering
\scalebox{0.85}{
\begin{tabular}{lcccc@{}}
\toprule
                & Relevance   & Correctness    & Accuracy      & Coverage \\ \midrule
\multicolumn{5}{c}{ChatGLM4 review}  \\ \midrule
QFS    & \textbf{2.98} & \textbf{5.97}  & \textbf{5.61}   & \textbf{5.49} \\
DBS       & 2.56 & 4.66 & 3.92 & 4.26     \\
\midrule
\multicolumn{5}{c}{Human review}  \\ \midrule
QFS    & \textbf{2.40} & \textbf{8.40}  & \textbf{4.30}   & \textbf{4.91} \\
DBS       & 1.80 & 5.71 & 3.70 & 4.12     \\
\bottomrule
\end{tabular}}
\caption{Human and LLM review results of query-focused summary and decomposition-based summary. QFS means query-focused summary, DBS means decomposition-based summary} 
\label{summary result}
\end{table}
\paragraph{Influence of posterior summary utilization.} Results in Table \ref{retrieval result} have already shown that Posterior Summary Utilization can greatly improve the performance of MDR, especially when reasoning is required. In this section, we further evaluate the difference between query-focused summary generation and decomposition-based summary generation originally used by MDR through reviewed by 3 human experts and ChatGLM4, a powerful multilingual LLM. Results are shown in Table \ref{summary result}. Our method achieves better performance. This result demonstrates the superiority of our method. Furthermore, the overall standard deviation of our method is 1.65, while that of decomposition-based summary generation is 2.86. This further proves the stability of our method.
\paragraph{Influence of momentum posterior regularization.} To evaluate the effectiveness of Momentum Posterior Regularization, we evaluate the 2nd hop retrieval performance using 1st hop golden summaries on HotpotQA dev set. The results are shown in Table \ref{golden summary}. As the results show, MoPo shows a great improvement over MDR baseline model, while PR even performs worse than our baseline model. This experiment confirms Momentum Posterior Regularization's effectiveness.
\begin{table}[t]
    \centering
    
    \begin{tabular}{lccc}
\toprule
    & \textbf{Bridge}           & \textbf{Comparison}  & \textbf{Total}            \\ \midrule
MDR    & 89.26 & 96.26 & 90.44  \\
PR & 77.46 & 97.20 & 80.67  \\ \midrule
MoPo & \textbf{93.74} & \textbf{99.13} & \textbf{94.67} \\ \bottomrule
\end{tabular}
\caption{Retrieval performance in Recall@100 on HotpotQA using 1st hop golden summary }
\label{golden summary}
\end{table}

\begin{table}[t]
\centering
\scalebox{0.9}{
\begin{tabular}{lcccc@{}}
\toprule
\textbf{Model}                & R@2   & R@20    & R@50      & R@100 \\ \midrule
MDR    & 94.38 & 94.85  & 95.76   & 96.49       \\
MDR \textit{w/} S       & 94.71 & 95.31  & 96.08   & 96.58       \\
PR & 94.49 & 95.22  & 95.91   & 96.51       \\
MoPo         & \textbf{94.71} & \textbf{95.34}  & \textbf{96.12}   & \textbf{96.81}    \\
\bottomrule
\end{tabular}}
\caption{Retrieval performance in recall at k retrieved passages with Contriever as embedding model on HotpotQA dev set}
\label{retrieval contriever}
\end{table}
\begin{table}[t]
\centering
\scalebox{0.9}{
\begin{tabular}{lcccc@{}}
\toprule
\textbf{Model}                & R@2   & R@20    & R@50      & R@100 \\ \midrule
e5-base    & 49.32 & 68.52  & 73.24   & 76.41       \\
e5-base-v2       & 55.26 & 75.92  & 80.55   & 83.62       \\

\bottomrule
\end{tabular}}
\caption{Retrieval performance in recall at k retrieved passages with different zero-shot embedding base models and using MDR framework on HotpotQA dev set}
\label{retrieval zero-shot}
\end{table}
\subsection{Experiment with Different Base Model}\label{appendix: qa}
We also evaluate the retrieval ability of different models with Contriever\cite{izacard2022unsuperviseddenseinformationretrieval} as base embedding model. Results are shown in Table \ref{retrieval contriever}. From the results. we can find that MoPo still has a better performance compared with other baseline models.

Moreover, we further evaluate the retrieval ability of different base embedding models using MDR framework in Table \ref{retrieval zero-shot}. The main difference between e5-base and e5-base-v2 in this experiment is that the pre-training data of e5-base-v2 contains HotpotQA while the other one does not. Compared with results in Table \ref{retrieval result}, the results show that multihop framework is more effective.

\subsection{Loss Trend Comparison with MDR}
\label{loss_lamda1}
    
    

 \begin{figure}[t]
     \centering
     \includegraphics[width=\linewidth]{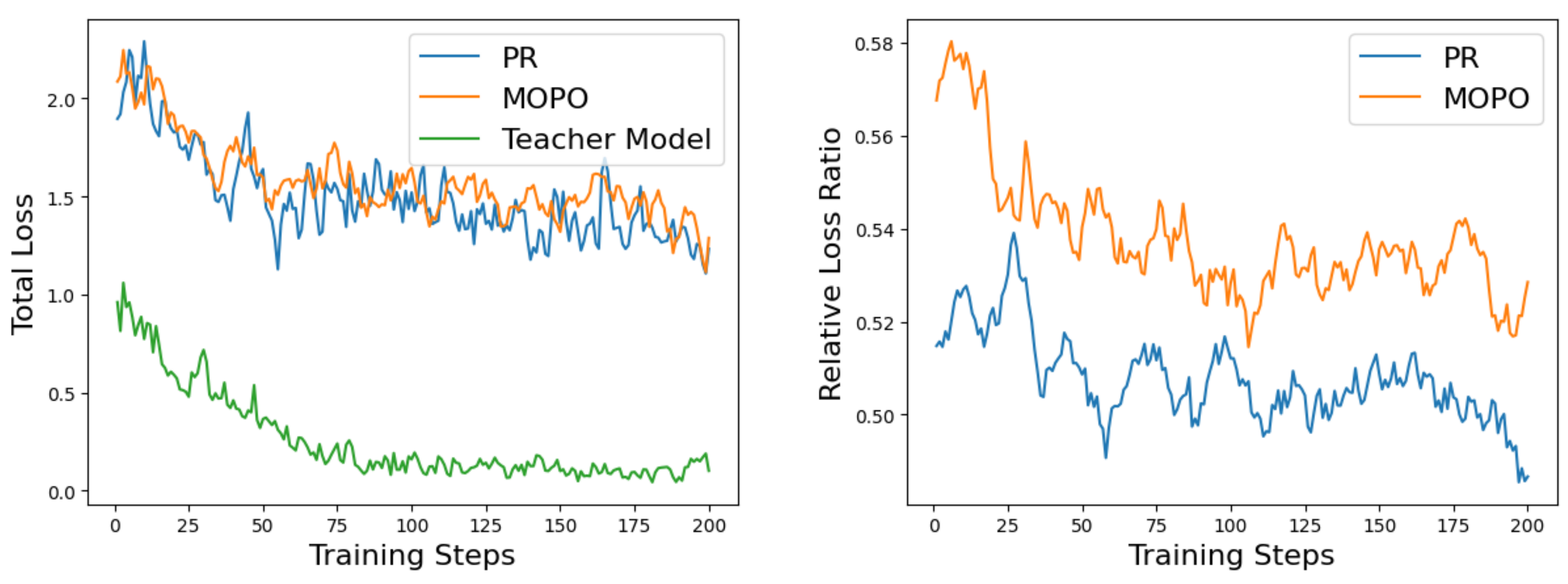}
 \caption{Total absolute and relative loss curve of PR and MoPo, $\lambda\myeq1.0$, where relative Loss Ratio = InfoNCE Loss / Total Loss. All curves are processed with the same smoothing factor.}
 \label{fig:loss_observation_1}
 \end{figure}
Following the analysis in Section \ref{loss analysis}. We further compared the loss curve of MoPo, {PR}$_{fixed}$. As shown in Figure \ref{fig:loss_observation_1}, teacher model converges much faster than other methods, which makes the training effect weakened.

\section{Selected Baseline Details}
\label{appendix:selected_baseline_details}
\paragraph{SelfAsk} \cite{press-etal-2023-selfask} exploits LLM to decide and generate next hop (follow-up) query, and uses a search engine for direct answer selection. 

\paragraph{IRCoT} \cite{trivedi-etal-2023-ircot} interleaves between CoT generation \cite{cot_wei_nips} and retrieving $K$ documents.

\paragraph{FLARE} \cite{jiang-etal-2023-flare} iteratively uses LLM to predict the upcoming sentences to anticipate future content and a retrieval to obtain relevant documents for sentence generation if it contains low-confidence tokens.

\paragraph{BeamAggr} \cite{chu-etal-2024-beamaggr} exploits LLM for question decomposition. For complex (multi-hop) questions, BeamAggr exploits a retrieval to get relevant documents for generating K (beamsize) answers for each question in the question tree.

\paragraph{MoPo+reranker+generation vs Reasoning with LLM} LLM-based methods are much more costly compared to our method due to the use of LLM for reranking, query formalization and answer generation. For example, IRCoT \cite{trivedi-etal-2023-ircot} also requires $L$ hops of iteration, each retrieves $K$ documents and generate the next CoT (question reformalization) or the final answer. However, our method exploits lightweight generation models for query reformalization and answer generation, whereas IRCoT requires resource-intensive LLMs to reason over a set of K retrieved documents. As a result, given the same dense retrieval at each iteration, the cost of IRCoT outweighs MoPo significantly. To reduce computation cost, IRCoT (and other models) often exploit a light-weight BM25 retrieval. This case, however, the QA performance is limited by the weak retrieval capability of BM25.

\end{document}